\documentclass[10pt,twocolumn,letterpaper]{article}

\usepackage[preprint]{cvpr}      
\usepackage{multirow}
\definecolor{cvprblue}{rgb}{0.21,0.49,0.74}
\usepackage[pagebackref,breaklinks,colorlinks,allcolors=cvprblue]{hyperref}

\def\paperID{114514} 
\def\confName{CVPR}
\def\confYear{2026}

\title{Rectifying Latent Space for Generative Single-Image Reflection Removal}

\author{Mingjia Li, Jin Hu, Hainuo Wang, Qiming Hu, Jiarui Wang, Xiaojie Guo$^{*}$\\
School of Software, Tianjin University, Tianjin, China\\
{\tt\small  \{mingjiali, jinhu, hainuo, wang\_jiarui, huqiming\}@tju.edu.cn}, \tt\small xj.max.guo@gmail.com \\
{\tt \small \url{https://gensirr.research.mingjia.li}}
}
  \newcommand{\blfootnote}[1]{%
    \begingroup
    \renewcommand\thefootnote{}\footnote{#1}%
    \addtocounter{footnote}{-1}%
    \endgroup
  }
  
\begin{document}
\twocolumn[{%
\renewcommand\twocolumn[1][]{#1}%
\maketitle
\vspace{-2.5em}
\includegraphics[width=\linewidth]{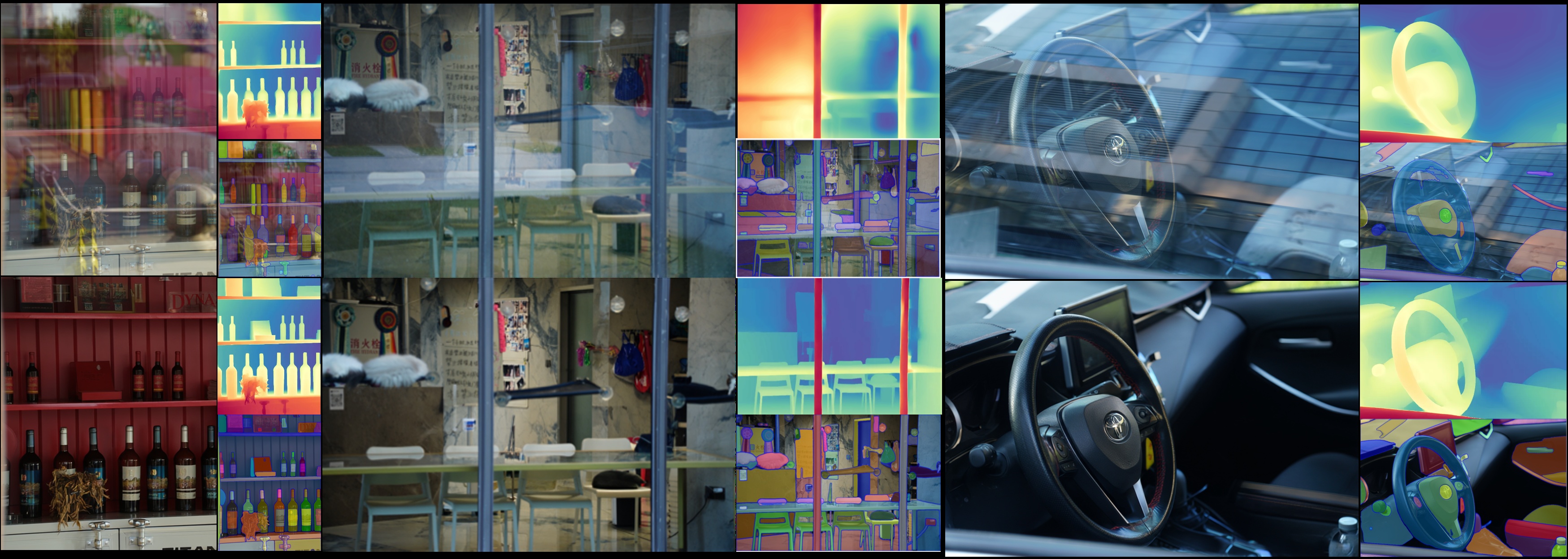}
\vspace{-20pt}
\captionof{figure}{\textit{Upper row} shows challenging real-world captures with strong reflections. Reflections interfere with the performance of downstream tasks, for instance, monocular depth estimation and zero-shot segmentation. \textit{Lower row} reveals that our proposed GenSIRR generalizes consistently well in these challenging cases, producing plausible and accurate results.}
\vspace{2em}
\label{fig:teaser}
}%
\vspace{-15pt}]

\blfootnote{*Corresponding Author.}

\begin{abstract}
Single-image reflection removal is a highly ill-posed problem, where existing methods struggle to reason about the composition of corrupted regions, causing them to fail at recovery and generalization in the wild. 
This work reframes an editing-purpose latent diffusion model to effectively perceive and process highly ambiguous, layered image inputs, yielding high-quality outputs. We argue that the challenge of this conversion stems from a critical yet overlooked issue, i.e., the latent space of semantic encoders lacks the inherent structure to interpret a composite image as a linear superposition of its constituent layers.
Our approach is built on three synergistic components, including a reflection-equivariant VAE that aligns the latent space with the linear physics of reflection formation, a learnable task-specific text embedding for precise guidance that bypasses ambiguous language, and a depth-guided early-branching sampling strategy to harness generative stochasticity for promising results. Extensive experiments reveal that our model achieves new SOTA performance on multiple benchmarks and generalizes well to challenging real-world cases. 

\end{abstract}

\section{Introduction}
\label{sec:intro}

When capturing photographs through transparent surfaces, a common phenomenon known as reflection superposition occurs. The transmission layer behind the surface, denoted as $B$, and the reflection layer on the surface, $R$, are superimposed on the sensor to form a composite image $I_{\text{obs}}$. A widely used model to describe this phenomenon is:
\begin{equation}
    I_{\text{obs}} = (1-\alpha)B + \alpha R,
\end{equation}
where $\alpha$ is a constant blending factor. Evidently, both the transmission layer $B$ and the reflection layer $R$ are natural images, rendering their mixture $I_{\text{obs}}$ semantically ambiguous. When our primary interest lies in the content behind the transparent surface, such as the interior of a vehicle, products in a display window as depicted in Figure~\ref{fig:teaser}, the presence of $R$ significantly obstructs our observation. This ambiguity caused by reflection superposition further degrades the performance of downstream tasks such as depth estimation, stereo matching, and segmentation \citep{tsin2003stereo,yang2016robust,costanzino2023learning,DBLP:conf/icra/LiuMGZZSZ25}, leading to confused estimations as illustrated in the lower part of the figure. As a long-standing challenge, this problem has been frequently addressed since the early days of computer vision \citep{pami/BergenBHP92,iccv/SchechnerKB98,cvpr/FaridA99,iccv/SarelI05,tog/AgrawalRNL05,iccv/SarelI05,cvpr/GuoCM14}. In the era of deep learning, the community has tackled this issue using a variety of models, from CNNs and Transformers to Diffusion models \cite{cvpr/ZhangNC18a,nips/HuG21,nips/hu2024single,hong2024differ,DBLP:journals/corr/abs-2503-17347}, leveraging various scales of pre-trained models to mitigate its inherent ill-posedness.

In this work, we highlight a critical yet largely overlooked elephant in the room: the composite image $I_{\text{obs}}$, due to its semantic ambiguity, cannot be well perceived by pre-trained models. Specifically, let us denote the encoder of a pre-trained model used for the Single Image Reflection Removal (SIRR) task as $E$. In the feature/latent space, the semantic embedding $z_I = E(I_{\text{obs}})$ is not equivalent to the linear combination of the embeddings of its constituent components ($z_B = E(B), z_R = E(R)$). That is, $z_I \neq (1-\alpha) z_B + \alpha z_R$. For pre-trained models driven by image classification tasks, this issue is less pronounced due to the widespread use of techniques like Mixup \cite{DBLP:conf/iclr/ZhangCDL18,DBLP:conf/iclr/UddinMSCB21}. However, for Latent Diffusion Models (LDMs) trained on image-text pairs \cite{DBLP:conf/nips/SchuhmannBVGWCC22}, this aspect has not been addressed. As shown in our experimental analysis in Table~\ref{tbl:vae_reconstruction_wrapped}, although a standard reconstruction loss can improve the fidelity between the pure VAE pass ($D(E(X))$) and the input image $X$, where $D$ is the VAE's decoder, its impact on the final reflection removal performance is minimal (SIRR on Real20). The significant improvement in reflection removal performance stems from the explicit alignment of $z_I$ with $(1-\alpha) z_B + \alpha z_R$. This indicates that making the pre-trained model $E$ understand that $I_{\text{obs}}$ is a superposition of $B$ and $R$, and explicitly regularizing this in the latent space, is remarkably beneficial for their separation. Foreseeably, this strategy can be extrapolated to other layer superposition problems, such as watermark removal, and alpha matting~\citep{DBLP:conf/iccv/NiuZZZ23,DBLP:conf/cvpr/XuPCH17}. Due to space constraints, this paper focuses on the task of single-image reflection removal.

For tackling the SIRR task with Latent Diffusion Models, the necessity of using natural language to describe the transmission/reflection scene has become a barrier to practical application. Given the inherent semantic ambiguity of the composite image $I_{\text{obs}}$, image tagging/captioning models~\citep{DBLP:conf/cvpr/ZhangHMLLXQLLLG22,DBLP:conf/icml/0001LXH22} struggle to automatically derive correct textual descriptions. To address this issue, we propose a Learnable Task-Specific Text Embedding (LTE). We initialize the LTE with the Fixed Text Embedding (FTE) derived from a constant text input and update the LTE as the model optimizes. Experiments show that our LTE can significantly enhance the performance of the LDM on the SIRR task.
\begin{table}[t]
    \centering
    \resizebox{\linewidth}{!}{%
    \begin{tabular}{@{}lccc@{}}
        \toprule
        \multirow{2}{*}{Model} & \multicolumn{1}{c}{Non-Mixed Rec.} & \multicolumn{1}{c}{50\% Mixed Rec.} & \multicolumn{1}{c}{SIRR on Real20} \\
        \cmidrule(lr){2-2} \cmidrule(lr){3-3} \cmidrule(lr){4-4}
        & SSIM$\uparrow$ & SSIM$\uparrow$ & SSIM$\uparrow$ \\
        \midrule
        FLUX-VAE & 0.909 & 0.859 & 0.841 \\
        $\mathcal{L}_{\mathrm{recon}}$ & 0.918 & 0.876 & 0.842 \\
        $\mathcal{L}_{\mathrm{recon}}+  \mathcal{L}_{\mathrm{equiv}}$ & \textbf{0.919} & \textbf{0.877} & \textbf{0.871} \\
        \bottomrule
    \end{tabular}%
    }
    \vspace*{-4pt}
    \caption{
        SSIM comparison on synthetic VAE reconstruction and real-world SIRR. Best results are in \textbf{bold}.
    }    \label{tbl:vae_reconstruction_wrapped}
    \vspace*{-10pt}
\end{table}

Another factor constraining the practical application of LDMs is the inherent stochasticity of generative models~\citep{DBLP:conf/nips/DhariwalN21}. During the denoising process with different initial noise samples, we can obtain output results with varying quality. At this juncture, if an evaluation metric could assess the quality of reflection removal at early timesteps of the denoising process and select the most promising sample to continue the subsequent denoising steps, it would be possible to achieve high-quality results from multiple random samples within a total time close to that of a single denoising run. We term this technique the Depth-guided Early-Branching Sampling strategy (DEBS).

 Our primary contributions are summarized as follows:
\begin{itemize}
    \item We propose GenSIRR, a novel pipeline that adapts the LDM for SIRR by introducing a reflection-equivariant VAE (re-VAE), which restructures the latent space to align with the linear physics of reflection formation.
    \item We design a Learnable Task-Specific Text Embedding (LTE) that bypasses the need for ambiguous natural language prompts, providing direct and optimized task-specific guidance to the generative model.
    \item We propose a depth-guided early-branching sampling strategy to secure faithful reflection removal. Our method sets a new state-of-the-art on public benchmarks, with superior generalization to challenging real-world images.
\end{itemize}

\section{Related Work}
\subsection{Single Image Reflection Removal}
\textit{Prior-based Methods.} Pioneering work on SIRR formulated the task as an optimization problem, regularized by handcrafted priors derived from the physical properties of the statistical regularities of natural images~\citep{LrvinPAMI2007}. These methods sought the most plausible decomposition by minimizing an objective function that balanced data fidelity with these priors. Prominent examples include relative smoothness assumptions~\citep{wacv/ChungCWC09, cvpr/LiB14}, where the transmission layer is expected to be smoother than the reflection, gradient sparsity~\citep{nips/LevinZW02, cvpr/LevinZW04, iccv/FanYHCW17}, and the detection of ghosting cues from double reflections~\citep{DBLP:conf/cvpr/ShihKDF15}. While insightful, they are constrained by a reliance on fragile assumptions, which are frequently violated in complex, real-world scenes. Consequently, the performance is often limited to controlled environments, exhibiting poor generalization to in-the-wild data~\citep{WANCVPR2018}. Nevertheless, these prior-based approaches established the conceptual groundwork for the field,  influencing the design of subsequent deep-learning methods.

\noindent\textit{Learning-based Methods.} The advent of deep learning~\citep{iclr/SimonyanZ14a, he2016deep}, particularly CNNs, marked a paradigm shift in SIRR~\citep{Zhang_2018_CVPR}. By training on synthetic datasets, these methods learn to perform the decomposition in an end-to-end fashion. The foundational insight was the use of semantic priors from pre-trained classification networks. Zhang \textit{et al.}~\citep{Zhang_2018_CVPR}, for instance, pioneered this by leveraging hypercolumn features from VGG-19~\citep{Hariharan_2015_CVPR} to imbue their model with greater semantic awareness, while ERRNet~\citep{Wei_2019_CVPR} further explored this by training with misaligned image pairs.
This core concept spurred a wave of architectural innovation, with increasingly sophisticated modeling of the relationship between the transmission and reflection layers. One strategy is a two-stage approach, where the network first estimates one component to guide the prediction of the other. RAGNet~\citep{apin/LiLYLRZ23} initially estimates the reflection to guide transmission recovery. A parallel line pursues simultaneous estimation through dual-stream networks. The YTMT strategy~\citep{nips/HuG21} exemplifies this by restoring both layers concurrently with an interactive module; however, its reliance on a linear physical assumption limited its performance. Other works, such as BDN~\citep{Yang_2018_ECCV} and IBCLN~\citep{Li_2020_CVPR}, employ iterative refinement, but their simpler interaction models can sometimes lead to heavy ghosting artifacts. More recent architectures have introduced even more complex interaction mechanisms. Dong \textit{et al.}~\citep{Dong_2021_ICCV} developed an iterative network that estimates a probabilistic reflection confidence map, while DSRNet~\citep{iccv/Hu23} introduced a mutually gated interaction mechanism. To capture long-range dependencies, sophisticated interactions, like the DSIT~\citep{DBLP:conf/nips/HuW024} and RDNet~\citep{cvpr/zhao2025reversible} have also been adapted, pushing the state-of-the-art. Recently, RRW~\citep{cvpr/zhu2024revisiting} proposed to detect the reflection area and then execute removal. 
Despite their success, they rely on semantic encoders (\textit{e.g.}, VGGNet~\cite{iclr/SimonyanZ14a} and Swin Transformer~\cite{DBLP:conf/iccv/LiuL00W0LG21}) that are inherently short of processing layer-superimposed inputs with ambiguous semantics. Moreover, due to lacking a powerful generative prior to synthesize missing information for regions heavily occluded by reflections, these models tend to leave strong residual artifacts.

\noindent \textit{Generative Methods.} The need to provide plausible textures for the restored layer and to inpaint content in heavily occluded regions motivated a shift towards leveraging generative priors.
Early explorations in this domain utilized Generative Adversarial Networks (GANs)~\citep{cvpr/ZhangNC18a,cvpr/WeiYFW019,nips/HuG21} to enhance the perceptual quality of the restored transmission layer. 
While GANs can produce sharp textures, they were often employed to impose an adversarial loss, serving more as a perceptual regularizer to enforce realism rather than a generator to plausibly fill in missing information. Diffusion models provide a more stable and controllable way for conditional generation, as demonstrated by the state-of-the-art performance in tasks like super-resolution~\citep{Fei_2023_CVPR} and inpainting~\citep{cvpr/RombachBLEO22}.
However, adapting these models for SIRR is non-trivial, as the task requires precise layer disentanglement rather than detail synthesis. Initial explorations into diffusion-based SIRR, such as L-DiffER~\citep{eccv/hong2024differ} and PromptRR~\citep{wang2024promptrr}, have leveraged language guidance as conditioning signals. This approach, however, faces a critical bottleneck: the ambiguity of describing the reflection task in text. Furthermore, these models are typically trained from scratch on task-specific datasets, limiting their access to the vast world knowledge embedded in large-scale, pre-trained foundation models. Another recent work, DAI~\citep{DBLP:journals/corr/abs-2503-17347} attempts to leverage a one-step diffusion prior with a newly-collected dataset to improve its performance in the wild.
Despite producing visually plausible results, these generative approaches still do not address the issue of unstructured latent space for composite images we identified.
Furthermore, one-step priors, while efficient, lack the generative strength to restore heavily occluded regions (see Figure~\ref{fig:real_occu}). Multi-step methods, on the other hand, are hampered by their reliance on ambiguous text prompts and their inherent stochasticity leads to outputs of inconsistent quality. Selecting the best result from multiple runs is often computationally infeasible, limiting their practical use.
\begin{figure*}[t]
    \centering
    \includegraphics[width=\linewidth]{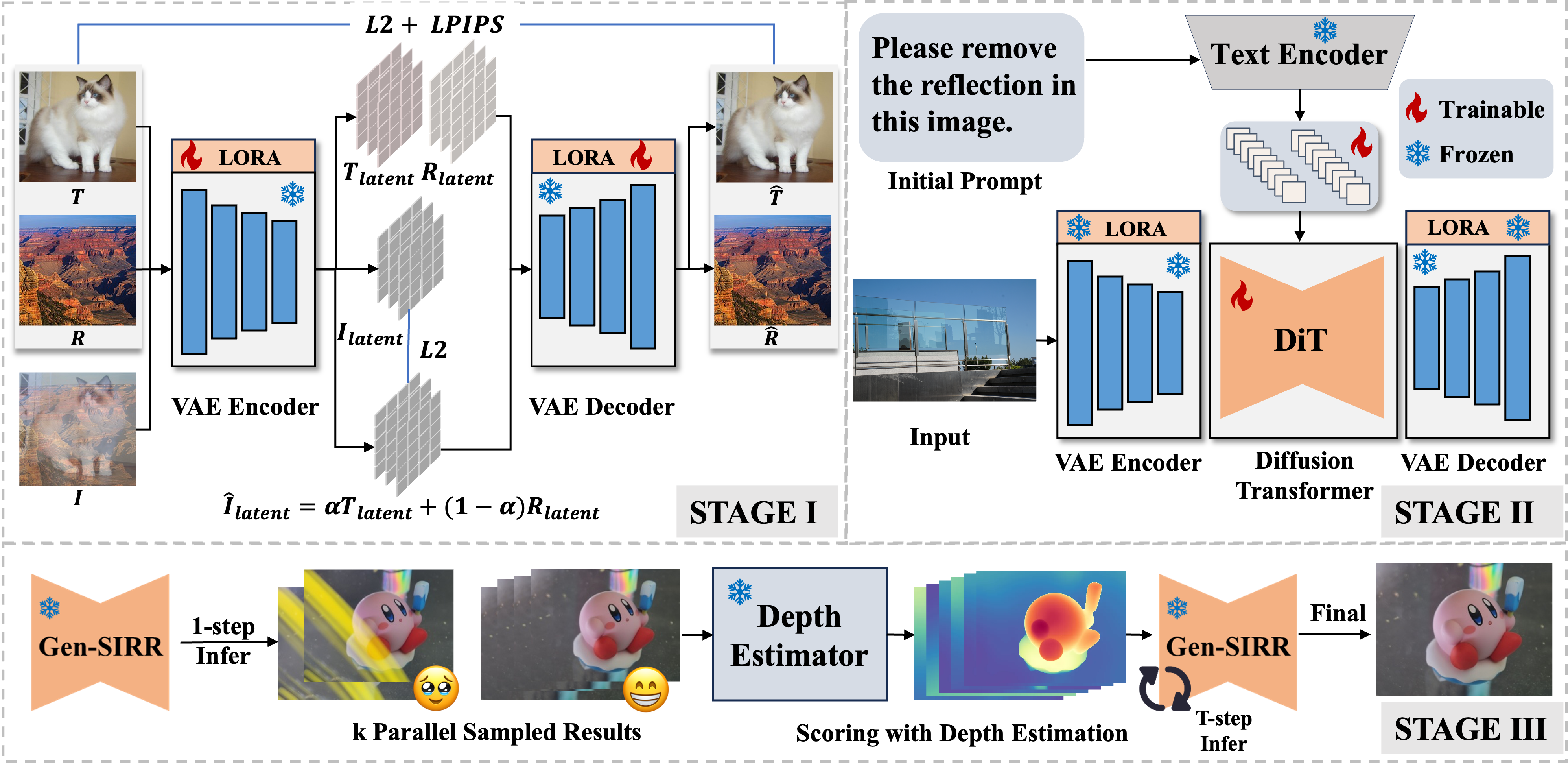}
    \caption{The overall pipeline of our proposed method. In Stage I, we use reflection-equivalence loss to regularize the latent space; during Stage II, the VAE encoder and decoder are frozen, while the text encoder encodes the initial prompt to initialize the learnable task embedding. In this stage, the task embedding and the DiT model are trained; Stage III is an additional test-time scaling stage. By sampling with multiple seeds, our scoring model can automatically select a faithful candidate with minimum computational overhead.}
    \label{fig:main}
\end{figure*}

\subsection{Generative Image Editing}
Our work is situated within the rapidly advancing field of generative image editing. While early methods often involved manipulating the latent space of GANs~\citep{DBLP:conf/nips/LingKLKTF21,DBLP:conf/cvpr/WangZFWC22,DBLP:conf/siggraph/PanTLLMT23}, the current state-of-the-art is now defined by large-scale text-to-image diffusion models like Stable Diffusion~\citep{cvpr/RombachBLEO22}, Imagen~\citep{DBLP:conf/nips/SahariaCSLWDGLA22}, and FLUX.1. The evolution of control mechanisms for these models has followed several key paradigms. The first and most common mode of control is the text prompt. A significant body of work, such as InstructPix2Pix~\citep{DBLP:conf/cvpr/BrooksHE23}, has focused on training models to follow editing instructions in natural language. While versatile for creative tasks, text is ill-suited for restoration tasks like SIRR, as describing the precise instructions of removing a reflection in words is both impractical and ambiguous. A second paradigm introduced explicit structural conditioning. Landmark methods like ControlNet~\citep{zhang2023adding} and T2I-Adapter~\citep{DBLP:conf/aaai/MouWXW0QS24} allow users to guide generation with auxiliary inputs like edge maps or depth maps. Works like OminiControl~\citep{DBLP:journals/corr/abs-2411-15098} and UNO~\citep{wu2025less} have explored fine-grained control by manipulating the model's internal features. While these provide powerful, general-purpose toolkits for manipulation, they often struggle with the high-fidelity content preservation required for restoration. Most recently, the field has seen the rise of massive, general-purpose image editing models like GPT-4o, FLUX.1 Kontext~\citep{DBLP:journals/corr/abs-2506-15742}, and Qwen3-Image-edit~\citep{wu2025qwenimagetechnicalreport}, which have demonstrated remarkable progress in instruction-following and content-preserving editing. However, their fidelity, while impressive, often falls short of the stringent requirements for high-precision image restoration tasks~\citep{DBLP:journals/corr/abs-2505-05621}. Furthermore, their reliance on the aforementioned paradigms still limits their applicability to tasks like SIRR. In this work, we leverage these powerful foundation models as a starting point and introduce the crucial missing piece for adapting them to the specialist task of SIRR.

\section{Methodology}

Our method is designed to tame a powerful pre-trained image editing LDM \footnote{Here, we choose FLUX.1 Kontext~\citep{DBLP:journals/corr/abs-2506-15742} as our base model.}for the precise and robust task of SIRR. To realize this, our pipeline is composed of two trainable components, as illustrated in Figure~\ref{fig:main}: (1) re-VAE that represents the reflection, transmission, and their mixture in a faithful manner, and (2) learnable task-specific text embeddings that instruct the LDM for the removal task. We also bring a depth-guided early-branching sampling strategy to improve the removal accuracy on extremely hard cases.

\subsection{Rectifying Latent Space for SIRR}

Our investigation begins with the VAE component of the LDM, which is for encoding the reflection-corrupted input image. As shown in Table~\ref{tbl:vae_reconstruction_wrapped}, while the original VAE excels at reconstructing standard images, its performance (measured by SSIM) degrades significantly on synthetic mixtures of background and reflection layers. Given encoder $E$ and decoder $D$, a straightforward solution is to finetune the VAE on these mixed images using a standard pixel-wise reconstruction loss, as suggested by stable diffusion~\citep{cvpr/RombachBLEO22}:
\begin{equation}
    \mathcal{L}_{\mathrm{recon}} = \Vert D(E(x)) - x \Vert_2^2 + \mathop{\mathrm{LPIPS}} (D(E(x)), x).  
\end{equation}
While this baseline approach improves the reconstruction metric on synthetic data by brute-forcing the VAE to accommodate the new distribution, the lack of structure proves critical when moving from reconstruction to reflection removal. Our goal is therefore not merely to improve reconstruction metrics, but to restructure the latent space to be \textit{reflection-equivariant}. We aim to align the VAE's latent geometry with the linear physics of reflection formation, where an observed image $I_{\text{obs}}$ can be seen as a linear blend of a background $B$ and a reflection $R$: $I_{\text{obs}} \approx (1-\alpha)B + \alpha R$. We enforce this property by finetuning the encoder $E$ into our proposed reflection-equivariant VAE (re-VAE), to penalize any deviation from the desired linear behavior in the latent space as:
\begin{equation}
    \mathcal{L}_{\mathrm{equiv}} = \left\| E(I_{\text{obs}}) - \left( (1-\alpha)E(B) + \alpha E(R) \right) \right\|_2^2.
\end{equation}
We term this as \textit{equivariance loss}, $\mathcal{L}_{\mathrm{equiv}}$, alongside the standard reconstruction loss. This combined objective compels the VAE to learn a latent space that is not only capable of faithfully reconstructing the input but is also explicitly aware of the linear superposition principle. During training, we sample a background $B$, a reflection $R$, and a random interpolation factor $\alpha \in [0, 1]$ to create the mixture $I_{\text{obs}}$. As shown in Table~\ref{tbl:vae_reconstruction_wrapped}, our method achieves reconstruction scores on synthetic data comparable to the baseline. The true benefit of our approach, however, is demonstrated in the SIRR performance. This proper latent structure is the key to its superior performance on real images, where the unstructured MSE-finetuned VAE fails to perform.
\begin{table*}[t]

\centering
\setlength{\tabcolsep}{10pt}
\begin{tabular}{llcccccccc}
    \toprule
    &\multirow{2}{*}{Methods} & \multicolumn{2}{c}{Real20 (20)} & \multicolumn{2}{c}{SIR2 (454)} & \multicolumn{2}{c}{Nature (20)} & \multicolumn{2}{c}{Average} \\
    \cmidrule(lr){3-4} \cmidrule(lr){5-6} \cmidrule(lr){7-8} \cmidrule(lr){9-10}
    & & PSNR & SSIM & PSNR & SSIM & PSNR & SSIM & PSNR & SSIM \\
    \midrule
    \multirow{8}{*}{\rotatebox{90}{Non-Gen.}} 
    & ERRNet(CVPR'19) & 22.89 & 0.803 & 23.55 & 0.882 & 22.18 & 0.756 & 23.47 & 0.874 \\
    & IBCLN(CVPR'20) & 21.86 & 0.762 & 24.20 & 0.884 & 23.57 & 0.783 & 24.08 & 0.875 \\
    & YTMT(NeurIPS'21)& 23.26 & 0.806 & 24.08 & 0.890 & 23.85 & 0.810 & 24.04 & 0.883 \\
    & Dong~\textit{et al.}(ICCV'21) & 23.34 & 0.812 & 24.25 & 0.901 & 23.45 & 0.808 & 24.18 & 0.894 \\
    & DSRNet(ICCV'23)& 23.91 & 0.818 & 25.71 & 0.906 & 25.22 & 0.832 & 25.62 & 0.899 \\
    & Zhu~\textit{et al.}(CVPR'24)& 21.83 & 0.801 & 25.48 & 0.897 & 26.04 & \underline{0.846} & 25.37 & 0.909 \\
    & DSIT(NeurIPS'24)& 25.22 & 0.836 & 26.43 & 0.911 & 26.77 & \textbf{0.847} & 26.40 & 0.905 \\
    & RDNet(CVPR'25) & 25.71 & 0.850 & 26.69 & 0.908 & 26.31 & \underline{0.846} & 26.63 & 0.903 \\
    \midrule
    \multirow{3}{*}{\rotatebox{90}{Gen.}} 
    & DAI~(AAAI'26) & 25.21 & 0.841 & 27.47 & 0.919 & 26.81 & 0.843 & 27.35 & 0.913 \\
    & Ours & \underline{27.27} & \underline{0.871} & \underline{27.99} & \underline{0.921} & \underline{27.30} & 0.838 & \underline{27.93} & \underline{0.916} \\
    & Ours + DEBS ($k=4$)& \textbf{27.58} & \textbf{0.881} & \textbf{28.08} & \textbf{0.937} & \textbf{27.34} & 0.840 & \textbf{28.03} & \textbf{0.931} \\
    \bottomrule
\end{tabular}
\caption{Quantitative comparison with SoTA methods on SIRR benchmarks (Real20, SIR2, Nature) and the average scores. Best results are in \textbf{bold} and second-best ones are \underline{underlined}. All the results are sampled from a fixed seed of 42. } 
\label{tab:quantitative_main}
\end{table*}

\begin{table*}[t]

\centering
\setlength{\tabcolsep}{6pt} 
\begin{tabular}{l ccccccccc ccc}
    \toprule
    Metric & Input & ERRNet & IBCLN & YTMT & Dong \textit{et al.} & DSRNet & DSIT & RDNet & DAI & Ours \\
    
    \midrule
    
    PSNR & \underline{26.64} & 22.60 & 24.33 & 22.20 & 23.76 & 23.30 & 24.77 &  24.87 &  25.27 & \textbf{27.76}\\
    
    SSIM &  \textbf{0.941}  & 0.802 & \underline{0.930} & 0.797 & 0.817 & 0.803 & 0.869 & 0.850 & 0.831 & 0.843 \\
    \midrule
     Masked PSNR & 22.09 & 21.59 & 22.62 & 21.87 & 20.94 & 22.11 & 23.35 &  23.87 &  \underline{24.03} & \textbf{27.19}\\
    
     Masked SSIM & 0.949  & 0.951 & 0.959 & 0.937 & 0.954 & 0.951 & 0.958 & 0.958 & \underline{0.972} & \textbf{0.984}\\
    \bottomrule
\end{tabular}
\caption{Comparison on OpenRR-val. Note that Input images score artificially high on standard metrics due to faint reflections. Masked metrics reveal the true performance gap. Best results are \textbf{bold}, second best \underline{underlined}. Please note that NONE of these methods, including ours, are trained on the OpenRR training set or selected on the OpenRR validation set.}
\vspace{-10pt}
\label{tab:openrr_horizontal}
\end{table*}
\subsection{Learnable Task Embedding for Guidance}

As stated in the Section~\ref{sec:intro}, adapting pre-trained Latent Diffusion Models for SIRR via standard text prompts is impractical~\citep{DBLP:conf/cvpr/ZhangHMLLXQLLLG22,DBLP:conf/icml/0001LXH22}. The inherent semantic ambiguity of the composite image $I_{\text{obs}}$ makes it nearly impossible for automated captioning models, or even humans, to provide a correct and useful textual description for guidance. To overcome this, we introduce a Learnable Task-Specific Text Embedding (LTE), denoted as $P_{\mathrm{task}}$, which are optimized to become a highly specialized, non-verbal task embedding. Our approach connects the model's text-based pre-training with our task-specific fine-tuning through a strategic initialization and optimization process. We begin initializing the embedding by tokenizing a simple, descriptive sentence: ``please remove the reflection within the image.'' The resulting sequence of text embeddings serves as initialization for our LTE. This initialization is crucial; it prompts the model's semantic pre-training, placing the optimization of the task-specific prompt in a semantically meaningful starting point of the embedding space, providing a strong inductive bias and a good starting point for optimization. These vectors are replacing the output of text encoders. During fine-tuning, the prompt vectors are untethered from their linguistic origins and are optimized via backpropagation. This allows $P_{\mathrm{task}}$ to evolve from a human-readable instruction into a direct, optimized task embedding that exists in the model's native operational space. In essence, this process distills the abstract concept of reflection removal into a set of precise, low-level instructions that are directly interpretable by the LDM's attention mechanisms.
The learnable prompt effectively becomes the operational command that instructs the LDM on \textit{how} to utilize the information provided by the input. This provides a clear and highly effective way to guide a large-scale generative model.

\subsection{Depth-guided Early-Branching Sampling}

\begin{figure}[t]
    \centering 
    \includegraphics[width=0.49\linewidth]{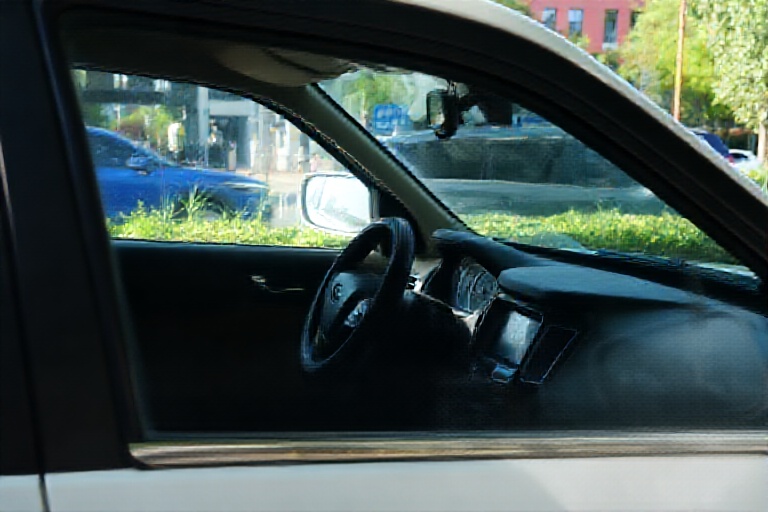}
    \includegraphics[width=0.49\linewidth]{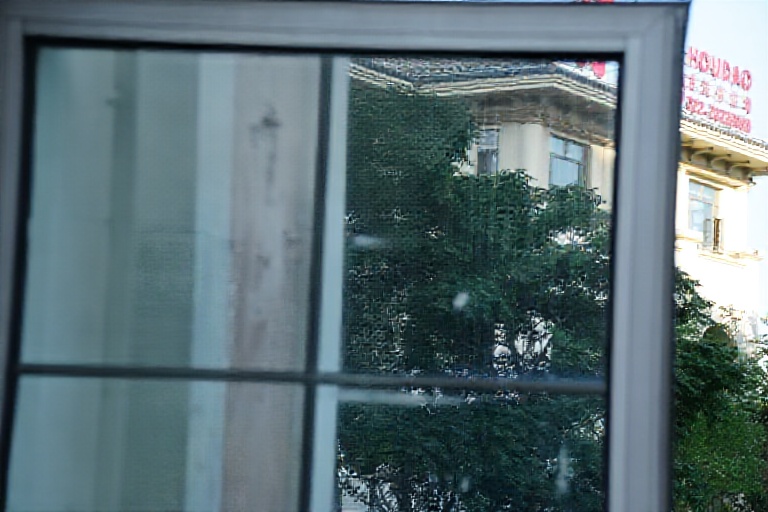}

    \includegraphics[width=0.49\linewidth]{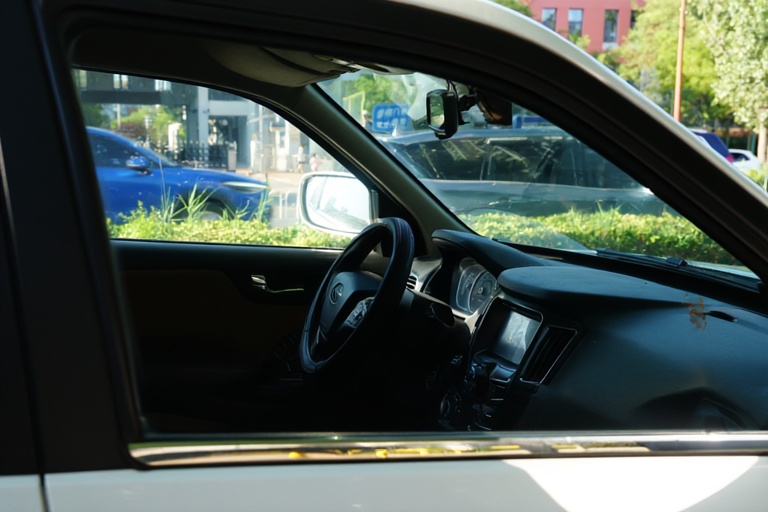}
    \includegraphics[width=0.49\linewidth]{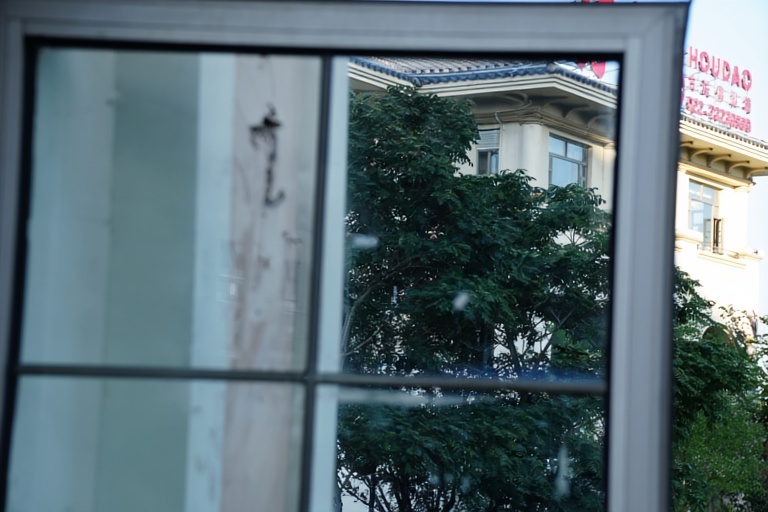}

    \caption{ One-step outputs (upper) and final results (lower).}
    \label{fig:depth_guide}
\end{figure}

While the stochasticity of generative models enables creative diversity, it poses a challenge for deterministic tasks like image restoration~\citep{dhariwal2021diffusion}. We reframe this randomness as an asset with our Depth-guided Early-Branching Sampling (DEBS) strategy. DEBS intelligently explores multiple restoration trajectories at a negligible computational cost to ensure a high-quality, deterministic output. Our key insight is that the structural success of reflection removal is determined within the very first denoising step, as demonstrated in Figure~\ref{fig:depth_guide}. We observe that if reflections are successfully suppressed in the single-step latent, they will remain absent in the final, fully denoised image. Based on this, DEBS initiates $k$ parallel denoising from different noise seeds for just one step.

Since we lack a ground truth during inference to pick the best candidate, we need a reference-free metric. Our key insight is that reflections disrupt the geometric consistency of a scene. A clean natural image typically possesses a coherent depth structure with piecewise smooth transitions. In contrast, residual reflections often manifest as "ghost" objects or semi-transparent layers that contradict the underlying scene geometry. When a monocular depth estimator is applied to an image with reflections, it often produces a noisy or ambiguous depth map, interpreting the reflection as a physical object floating in the foreground. 
We hypothesize that the candidate image yielding the \textit{deepest} depth map corresponds to the highest quality reflection removal. Since reflections act as foreground occlusions, their successful removal should reveal the true, more distant background, resulting in a global increase in estimated depth. By using a lightweight depth estimatorr~\cite{DBLP:conf/nips/YangKH0XFZ24} as a proxy for perceptual quality, we can effectively filter out suboptimal generation paths early in the sampling process. The candidate is then selected to proceed through the full $T$-step denoising process, while the other trajectories are discarded. This early-branching harnesses the exploratory power of multi-sampling to robustly guide the generation towards a superior output, with a computational overhead that is only marginally higher than a single inference pass.

\section{Experimental Validation}

\subsection{Implementation Details}
Our framework is implemented in PyTorch and consists of a two-stage training process: training the reflection-equivariant VAE, and then fine-tuning the LDM.

\begin{figure}[t]
    \centering
    \includegraphics[width=0.31\linewidth]{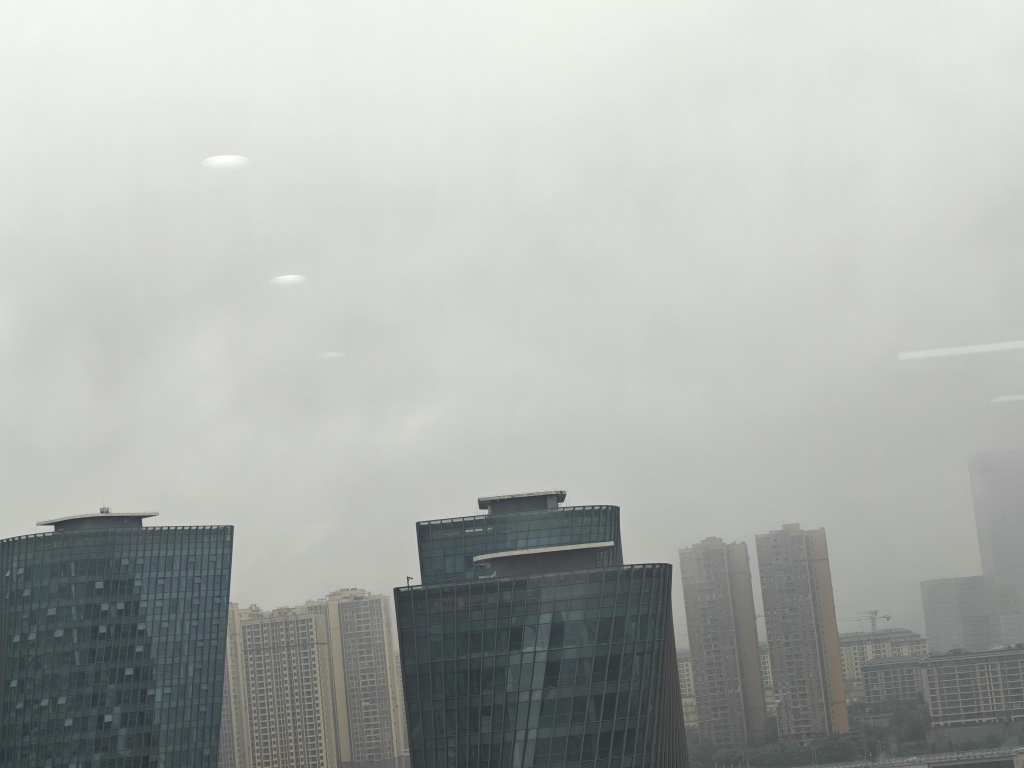}
    \includegraphics[width=0.31\linewidth]{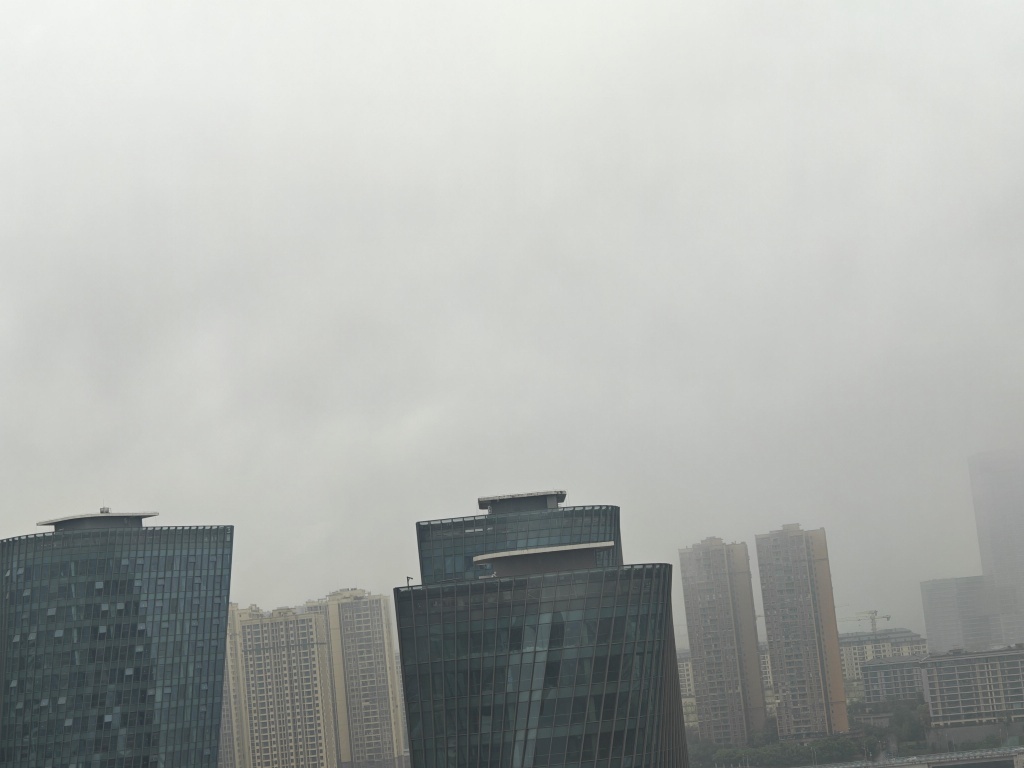}
    \includegraphics[width=0.31\linewidth]{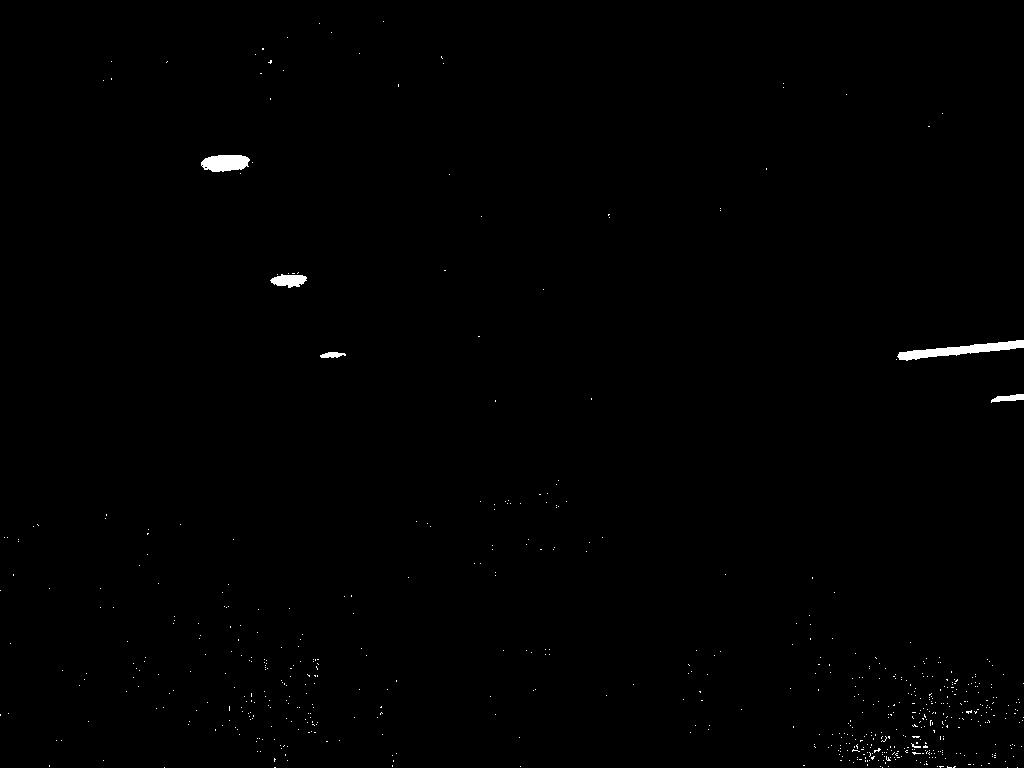}

    \includegraphics[width=0.31\linewidth]{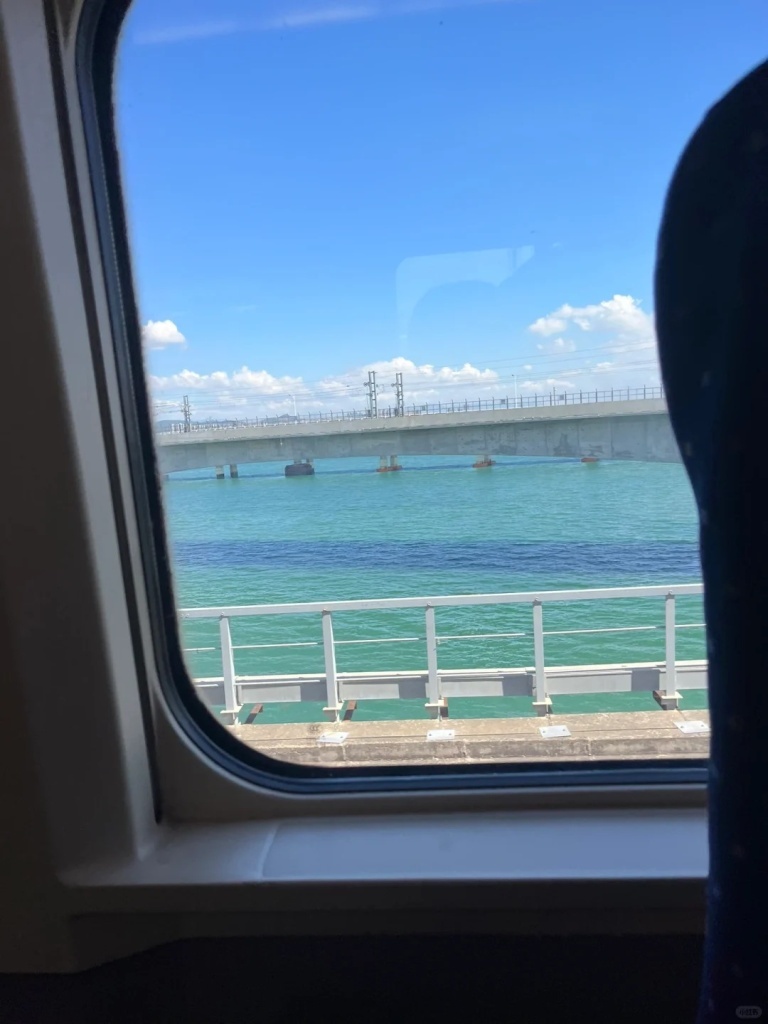}
    \includegraphics[width=0.31\linewidth]{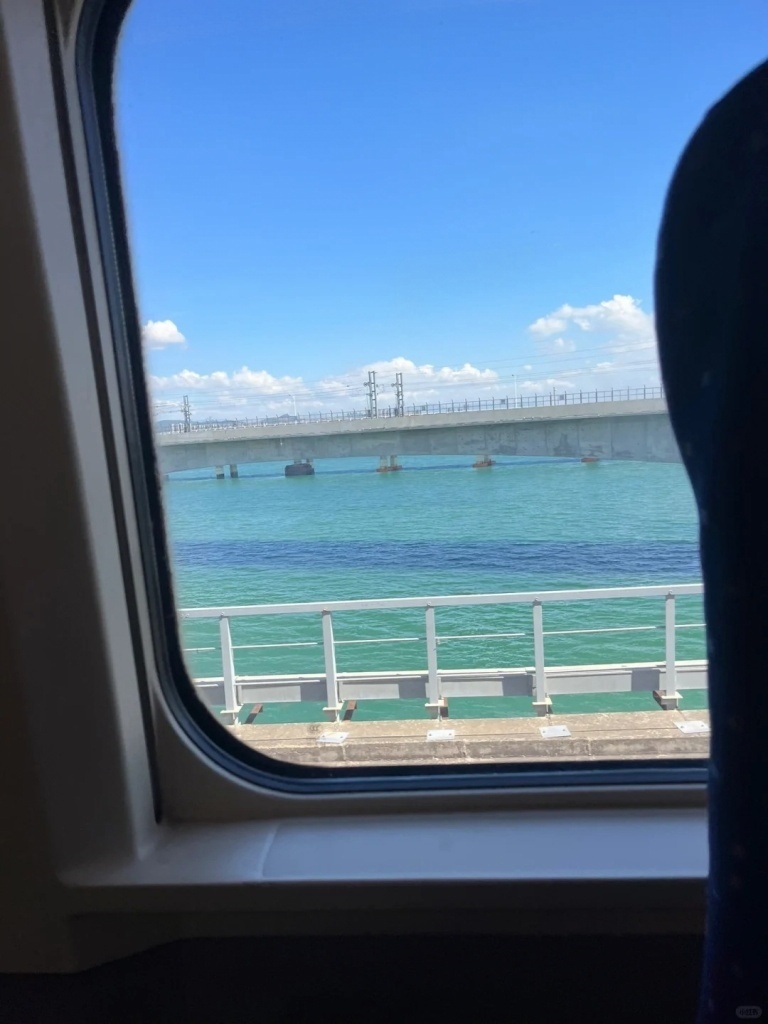}
    \includegraphics[width=0.31\linewidth]{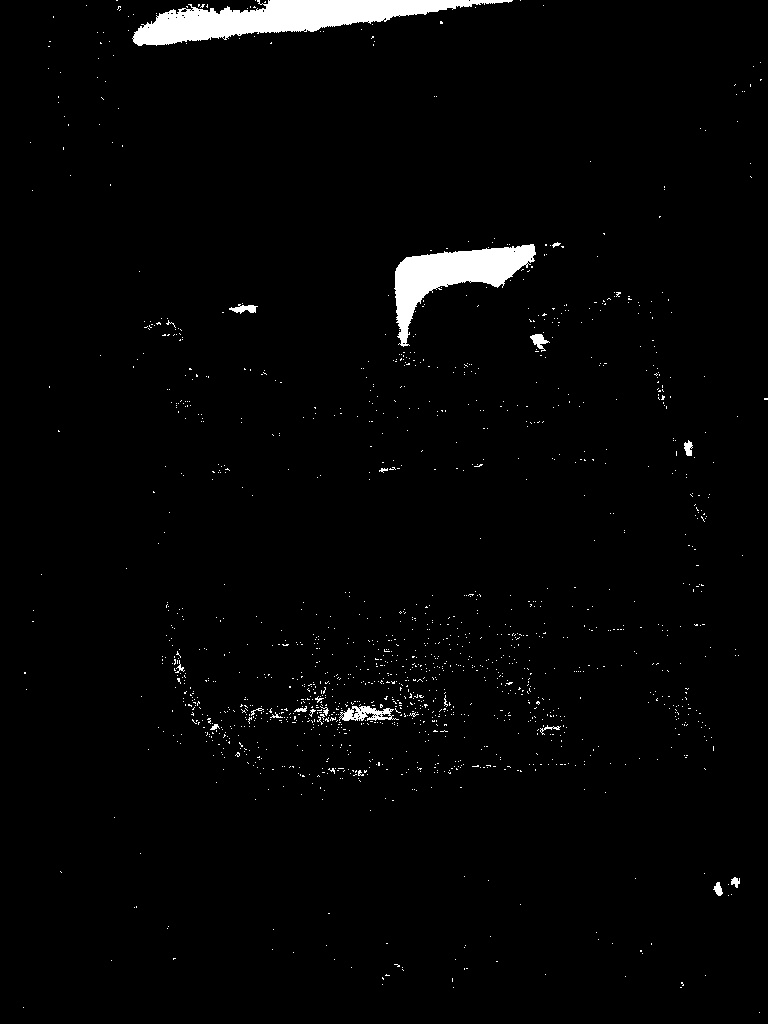}
    
   \begin{subfigure}{0.31\linewidth}
        \centering
        \subcaption{Input}
    \end{subfigure}
    \begin{subfigure}{0.31\linewidth}
        \centering
        \subcaption{GT}
    \end{subfigure}
    \begin{subfigure}{0.31\linewidth}
        \centering
        \subcaption{Mask}
    \end{subfigure}    
    \caption{Examples of input, ground truth, and the corresponding mask on the OpenRR-val dataset with a mask threshold of 5.}
    \label{fig:openrr_ma}
\end{figure}

\begin{figure*}[t]
    \centering 
    \includegraphics[width=0.24\textwidth,height=0.160\textwidth]{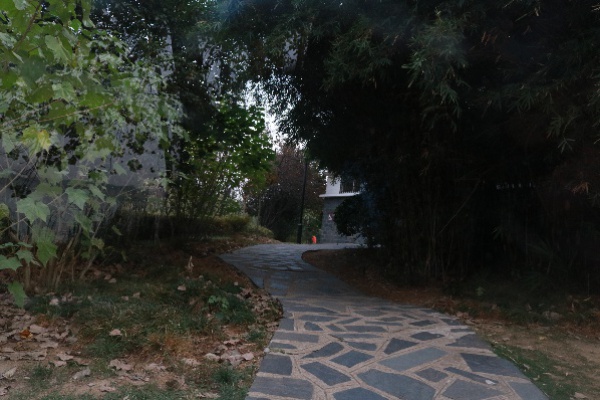}
    \includegraphics[width=0.24\textwidth,height=0.160\textwidth]{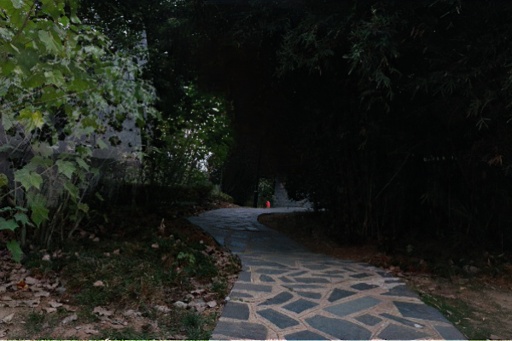}
    \includegraphics[width=0.24\textwidth,height=0.160\textwidth]{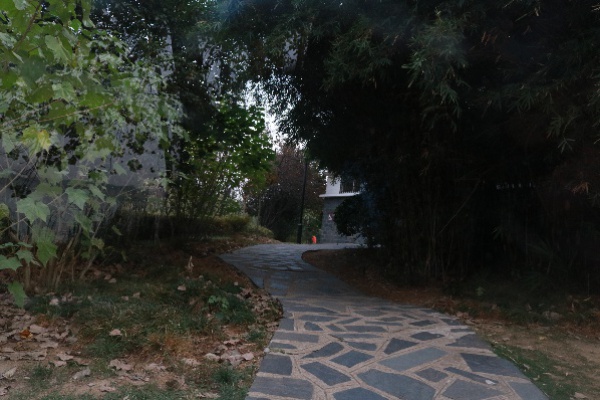}
    \includegraphics[width=0.24\textwidth,height=0.160\textwidth]{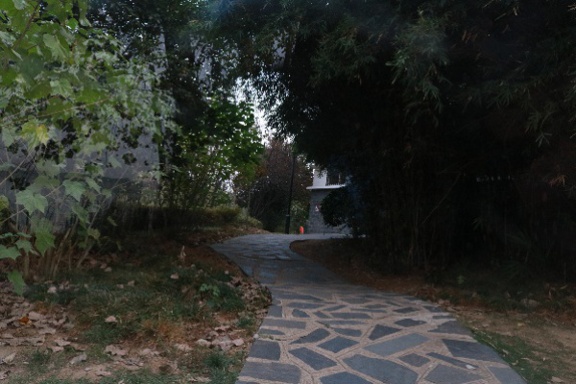}
    
    \includegraphics[width=0.24\textwidth,height=0.160\textwidth]{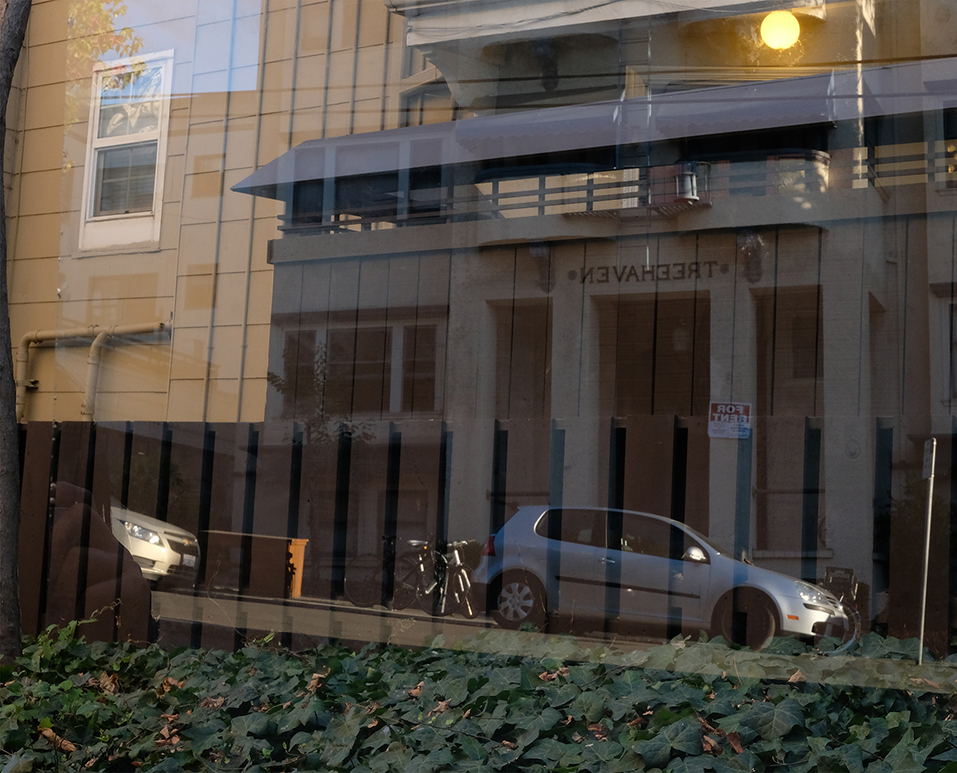}
    \includegraphics[width=0.24\textwidth,height=0.160\textwidth]{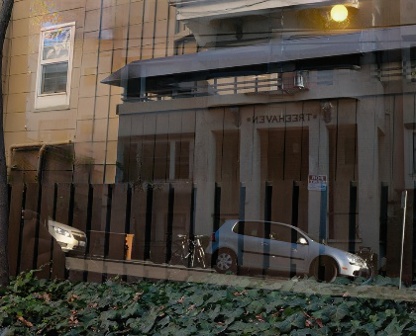}
    \includegraphics[width=0.24\textwidth,height=0.160\textwidth]{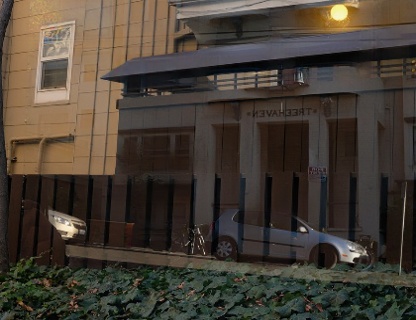}
    \includegraphics[width=0.24\textwidth,height=0.160\textwidth]{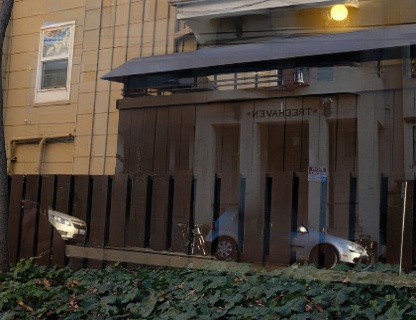}
    
   \begin{subfigure}{0.24\textwidth}
        \centering
        \subcaption{Input}
        \label{subfig:input}
    \end{subfigure}
    \begin{subfigure}{0.24\textwidth}
        \centering
        \subcaption{YTMT}
        \label{subfig:ERRNet}
    \end{subfigure}
    \begin{subfigure}{0.24\textwidth}
        \centering
        \subcaption{DSRNet}
        \label{subfig:IBCLN}
    \end{subfigure}
    \begin{subfigure}{0.24\textwidth}
        \centering
        \subcaption{DSIT}
        \label{subfig:Dong}
    \end{subfigure}

    \includegraphics[width=0.24\textwidth,height=0.160\textwidth]{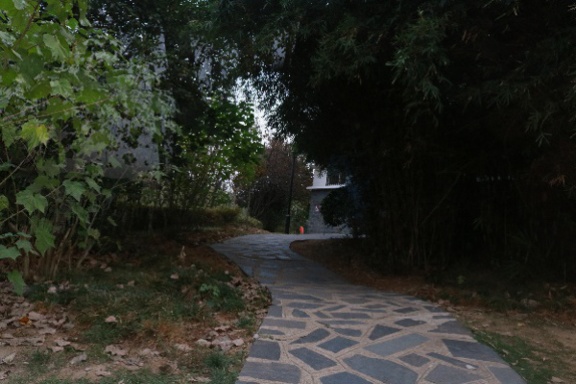}
    \includegraphics[width=0.24\textwidth,height=0.160\textwidth]{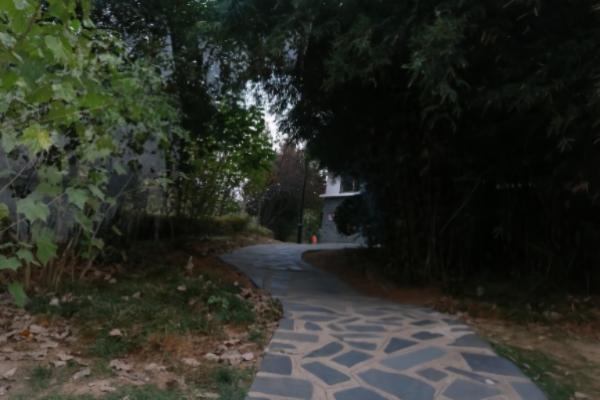}
    \includegraphics[width=0.24\textwidth,height=0.160\textwidth]{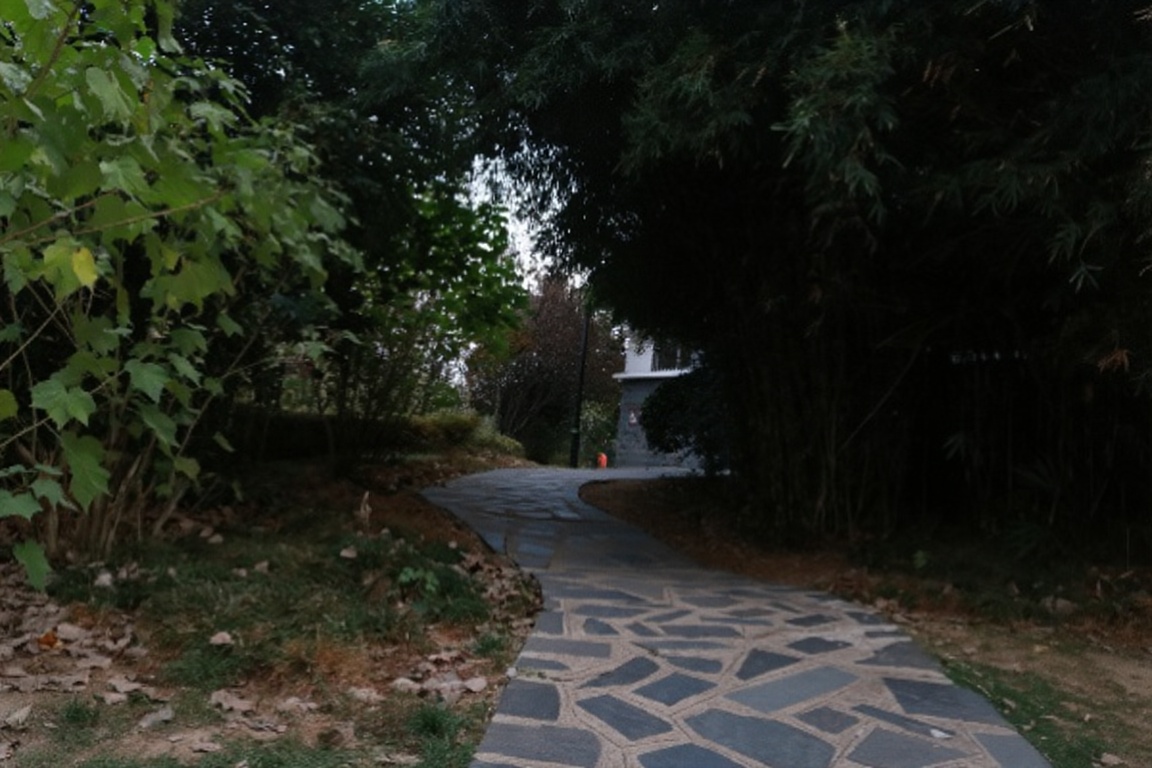}
    \includegraphics[width=0.24\textwidth,height=0.160\textwidth]{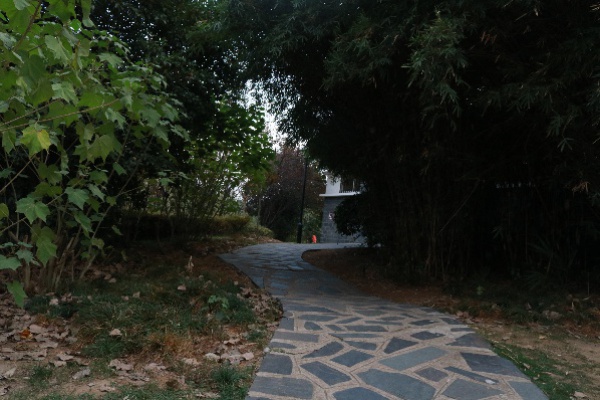}

    \includegraphics[width=0.24\textwidth,height=0.160\textwidth]{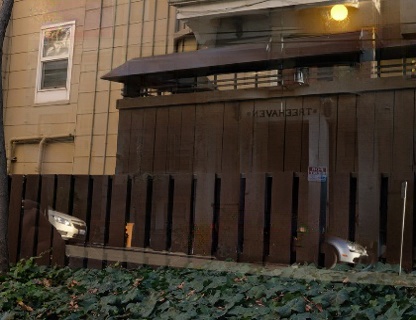}
    \includegraphics[width=0.24\textwidth,height=0.160\textwidth]{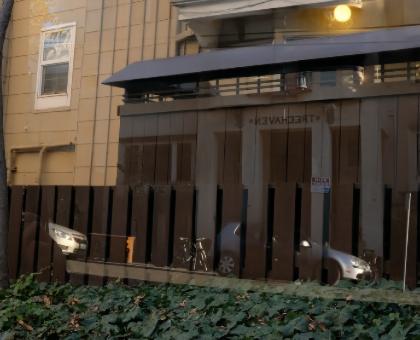}
    \includegraphics[width=0.24\textwidth,height=0.160\textwidth]{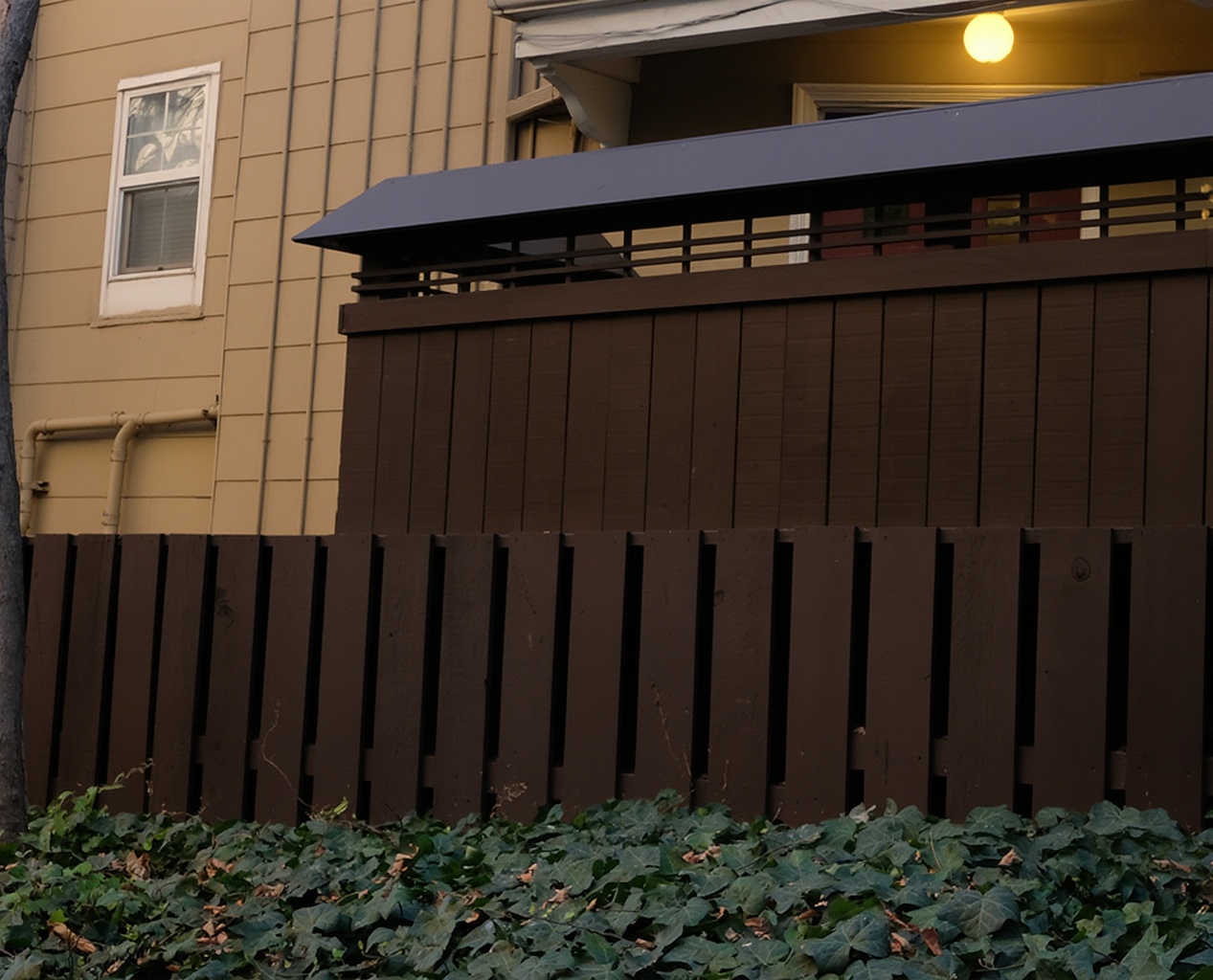}
    \includegraphics[width=0.24\textwidth,height=0.160\textwidth]{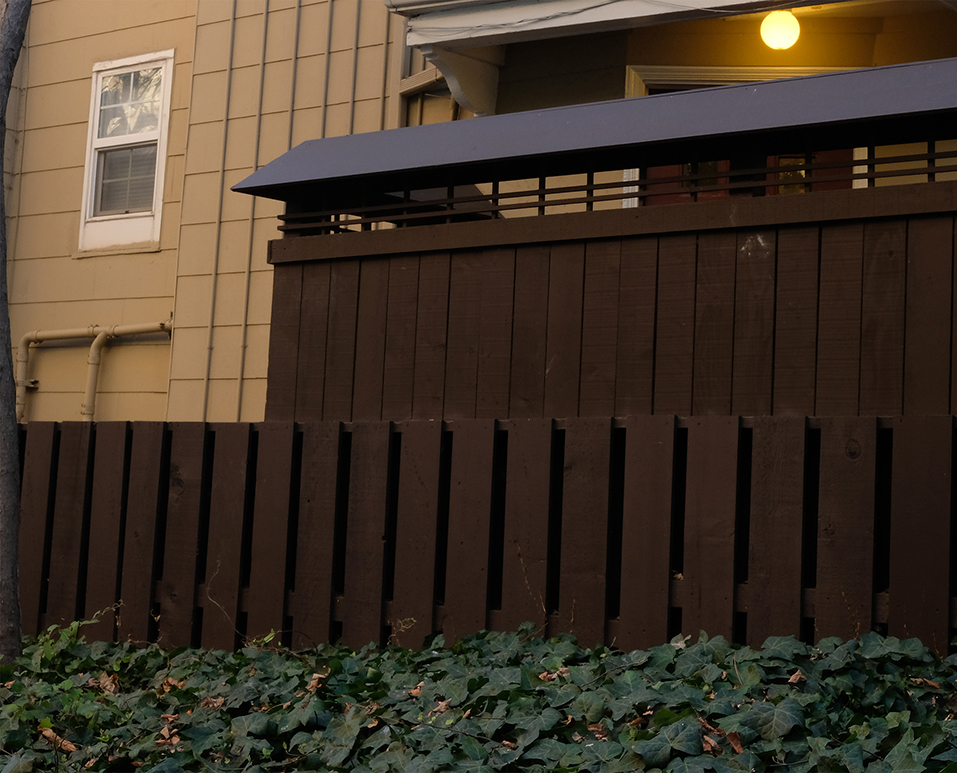}
    
   \begin{subfigure}{0.24\textwidth}
        \centering
        \subcaption{RDNet}
        \label{subfig:YTMT}
    \end{subfigure}
    \begin{subfigure}{0.24\textwidth}
        \centering
        \subcaption{DAI}
        \label{subfig:DSRNet}
    \end{subfigure}
    \begin{subfigure}{0.24\textwidth}
        \centering
        \subcaption{Ours}
        \label{subfig:zhu}
    \end{subfigure}
    \begin{subfigure}{0.24\textwidth}
        \centering
        \subcaption{GT}
        \label{subfig:input}
    \end{subfigure}
    \caption{Qualitative comparisons on Nature (top row) and Real20 (bottom row) datasets. Please zoom in for more details.}
    \label{fig:dataset}
\end{figure*}

\noindent\textit{Reflection-Equivariant VAE Training.}
The first stage focuses on training our VAE to learn a structured latent space for reflection. To make sure that the latent space won't change dramatically, we train a Low-Rank Adaptation (LoRA) adapter with a rank of 8. The training is conducted for 30,000 iterations using the AdamW optimizer with a learning rate of 1e-4. To learn a robust and general representation of reflections, we use the high-quality PD-12M~\citep{meyer2024publicdomain12mhighly} dataset, exposing the model to approximately 3.84 million unique images with a global batch size of 128.

\noindent\textit{DiT Fine-Tuning.}
The second stage involves fine-tuning the main reflection removal model. We initialize our model from a pre-trained FLUX.1 Kontext checkpoint to leverage its powerful generative prior and extensive world knowledge. The full model is then fine-tuned using the AdamW optimizer with a fixed learning rate of 1e-5 and a batch size of 32. Following established best practices~\cite{cvpr/zhao2025reversible, DBLP:conf/nips/HuW024} in the field, our training data is a curated mixture of real-world (Real~\cite{Zhang_2018_CVPR}, Nature~\cite{Dong_2021_ICCV}, RRW~{\cite{cvpr/zhu2024revisiting}) and synthetic images.

\subsection{Quantitative Performance Evaluation}

\noindent\textit{Datasets and Evaluation Metrics.} Our evaluation is on four widely-used benchmarks: \textit{Nature}~\citep{Dong_2021_ICCV}, \textit{SIR2}~\citep{wan2022sir2+}, \textit{Real}~\citep{Zhang_2018_CVPR}, and \textit{OpenRR}~\citep{Yang_2025_CVPR}.
We use the two most common metrics in SIRR to measure the quality of the recovered transmission layer, \emph{i.e.}, Peak Signal-to-Noise Ratio (PSNR) and Structural Similarity Index Measure (SSIM).

\noindent\textit{Baselines.} We compare GenSIRR against a representative set of recent and influential SIRR methods, including the CNN-based ERRNet~\citep{cvpr/WenT0LHH19}, IBCLN~\citep{Li_2020_CVPR}, YTMT~\citep{nips/HuG21} and Dong \emph{et al.}~\citep{Dong_2021_ICCV} as well as more recent architectures like DSIT~\citep{DBLP:conf/nips/HuW024} and RDNet~\citep{cvpr/zhao2025reversible}. We also included DAI~\citep{DBLP:journals/corr/abs-2503-17347}, a model trained with a more sophisticated dataset and large-scale pretraining. These baselines cover a range of architectural designs and reflect the current state-of-the-art.

\noindent\textit{Results on Benchmark Datasets.} As shown in Table~\ref{tab:quantitative_main}, GenSIRR establishes a new state-of-the-art across all benchmark datasets. Our method consistently outperforms all baseline models in both PSNR and SSIM, often by a significant margin. The performance gains are particularly pronounced on the more challenging datasets, which contain complex structures and diverse reflection types. This demonstrates the superior capability of our generative approach to handle difficult cases where others struggled. 

A critical observation regarding the OpenRR-val benchmark motivates our additional evaluation strategy. On this dataset, we observed that the unprocessed, reflection-corrupted input images often achieve the highest SSIM and the second-highest PSNR scores when compared directly against the ground truth (see Table~\ref{tab:openrr_horizontal}). This anomaly occurs because many reflections in this dataset are faint or localized; consequently, the pixel-wise difference between the input and the ground truth is minimal. Standard global metrics, therefore, fail to penalize distracting reflections and cannot reliably measure the perceptual success of the removal. To evaluate performance accurately, we employed mask-based evaluation. Inspired by RRW~\cite{cvpr/zhu2024revisiting}, we compute the difference map between the input and the ground truth to generate a reflection mask (threshold $> 5/255$). We then calculate metrics specifically within these masked regions. Figure~\ref{fig:openrr_ma} visualizes the input, ground truth, and evaluation masks. As shown in Table~\ref{tab:openrr_horizontal}, while global metrics favor the Input, our GenSIRR yields a clear advantage in the Masked PSNR and SSIM, demonstrating the superiority of our method in effectively removing reflection artifacts.

\begin{figure*}[t]
    \centering 
    \includegraphics[width=0.24\textwidth]{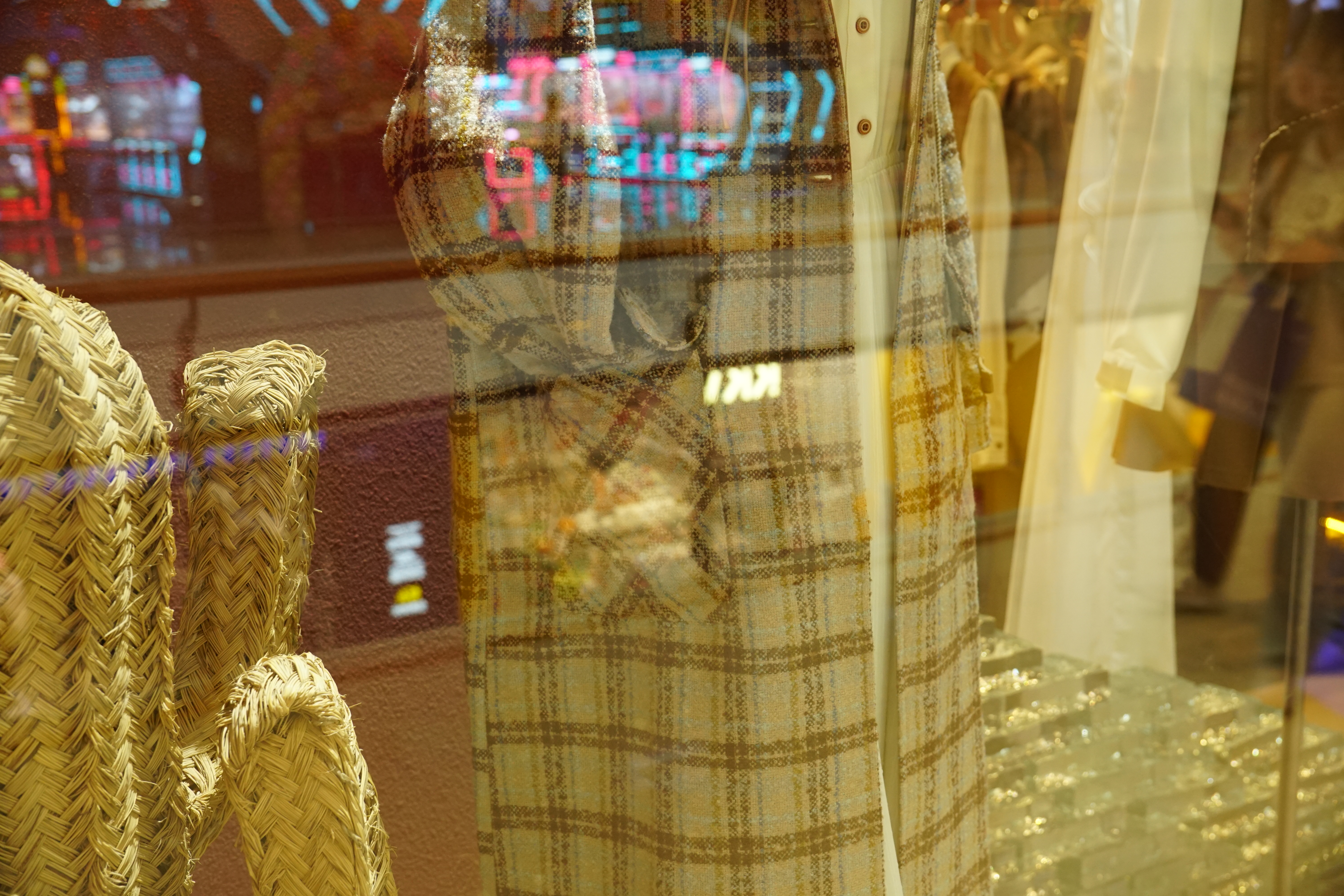}
    \includegraphics[width=0.24\textwidth]{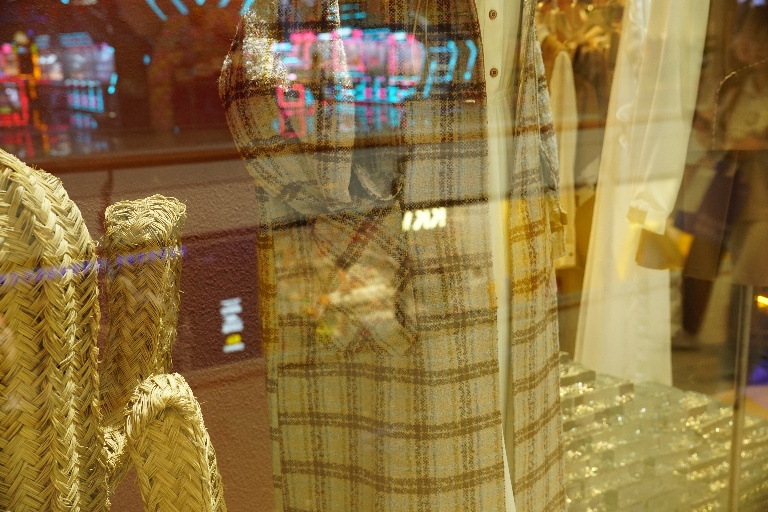}
    \includegraphics[width=0.24\textwidth]{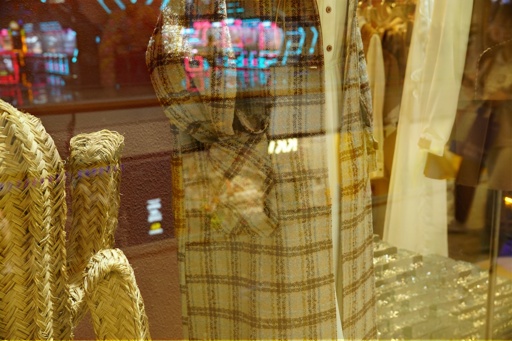}
    \includegraphics[width=0.24\textwidth,height=0.160\textwidth]{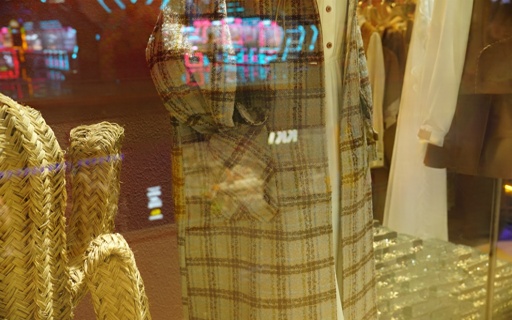}
    
    \includegraphics[width=0.24\textwidth]{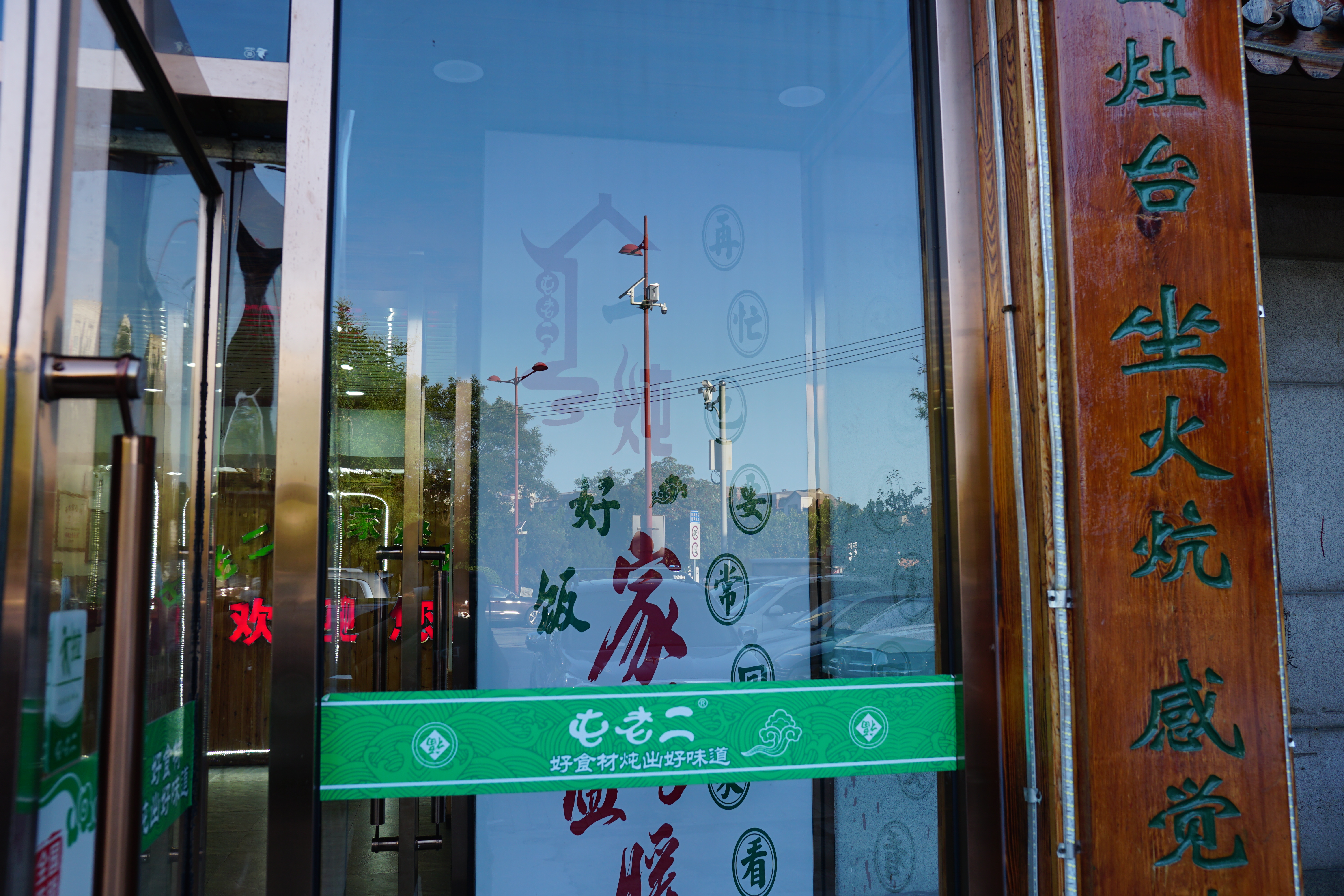}
    \includegraphics[width=0.24\textwidth]{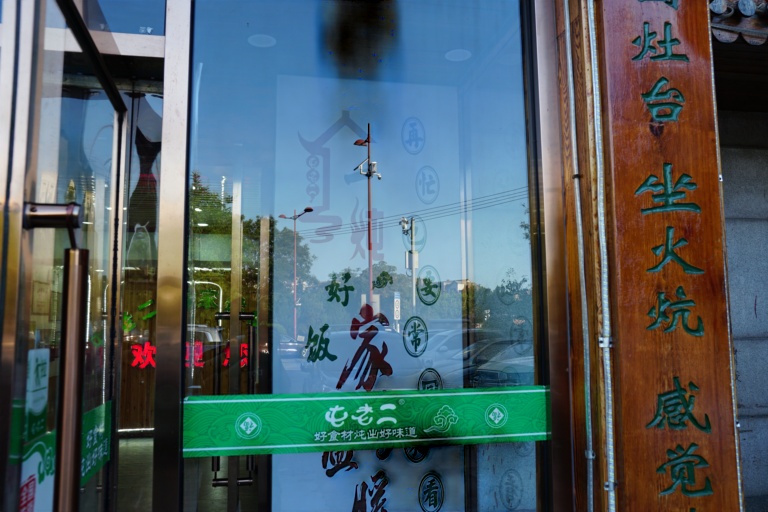}
    \includegraphics[width=0.24\textwidth]{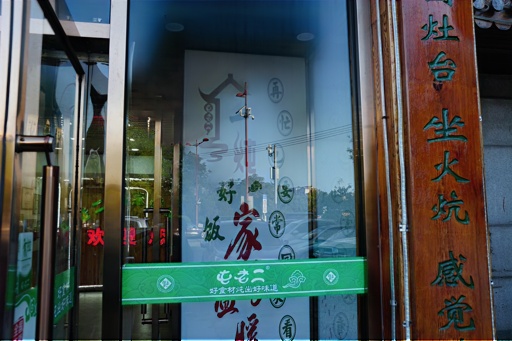}
    \includegraphics[width=0.24\textwidth, height=0.160\textwidth]{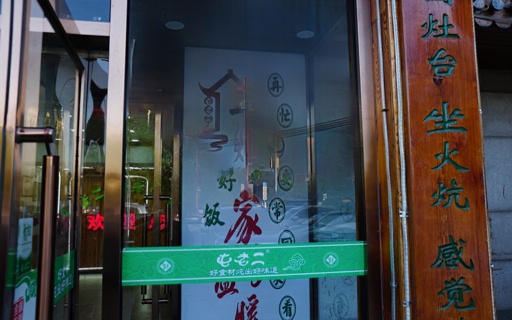}
    
   \begin{subfigure}{0.24\textwidth}
        \centering
        \subcaption{Input}
        \label{subfig:input}
    \end{subfigure}
    \begin{subfigure}{0.24\textwidth}
        \centering
        \subcaption{IBCLN}
        \label{subfig:ERRNet}
    \end{subfigure}
    \begin{subfigure}{0.24\textwidth}
        \centering
        \subcaption{YTMT}
        \label{subfig:IBCLN}
    \end{subfigure}
    \begin{subfigure}{0.24\textwidth}
        \centering
        \subcaption{DSRNet}
        \label{subfig:Dong}
    \end{subfigure}

    \includegraphics[width=0.24\textwidth]{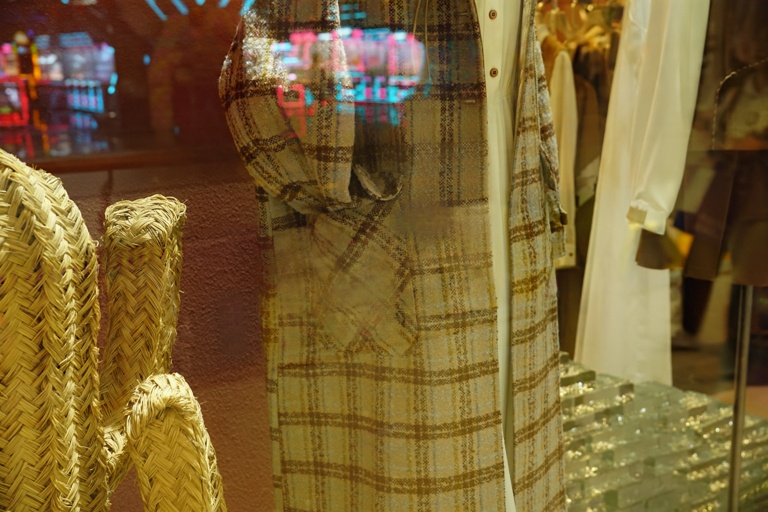}
    \includegraphics[width=0.24\textwidth]{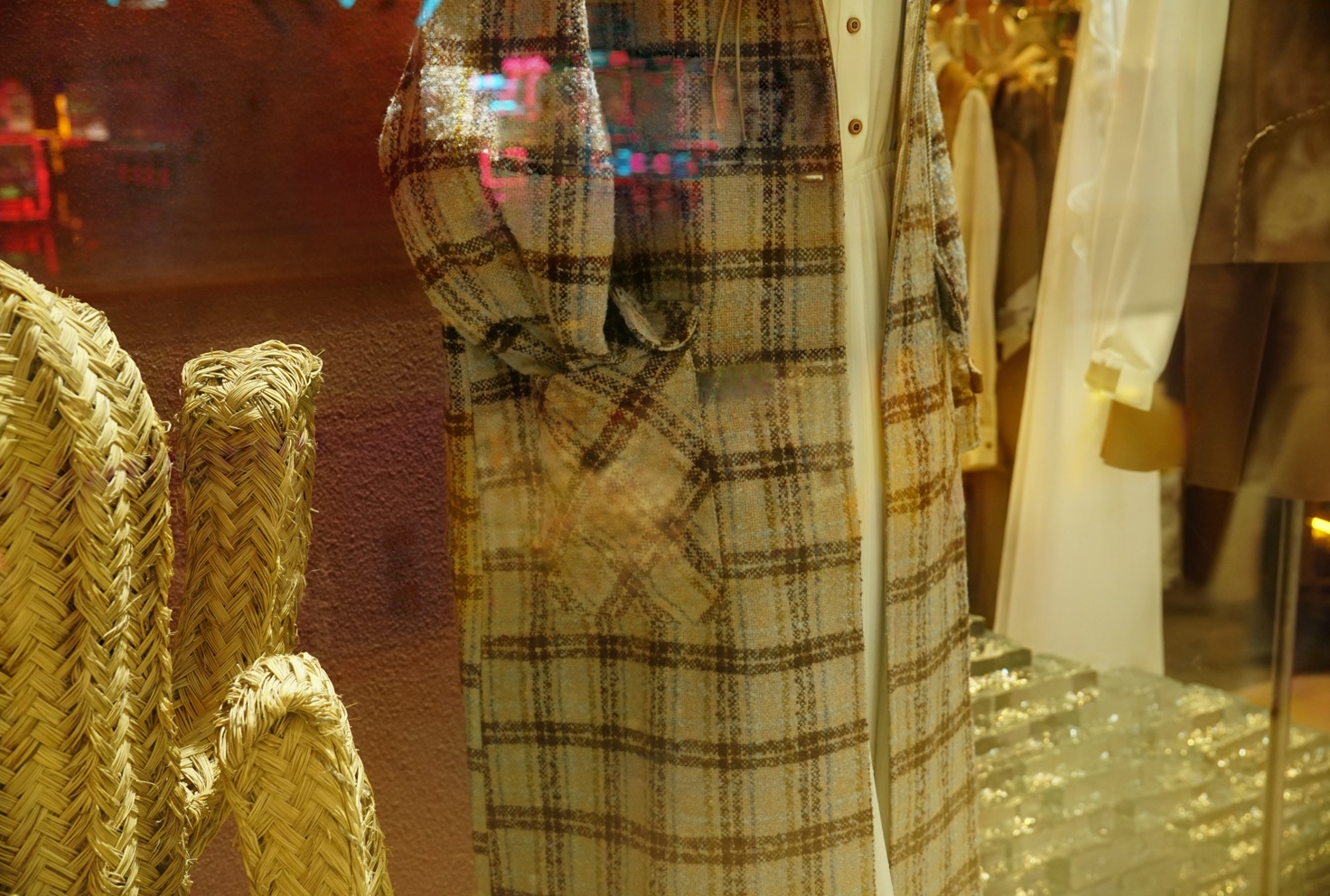}
    \includegraphics[width=0.24\textwidth]{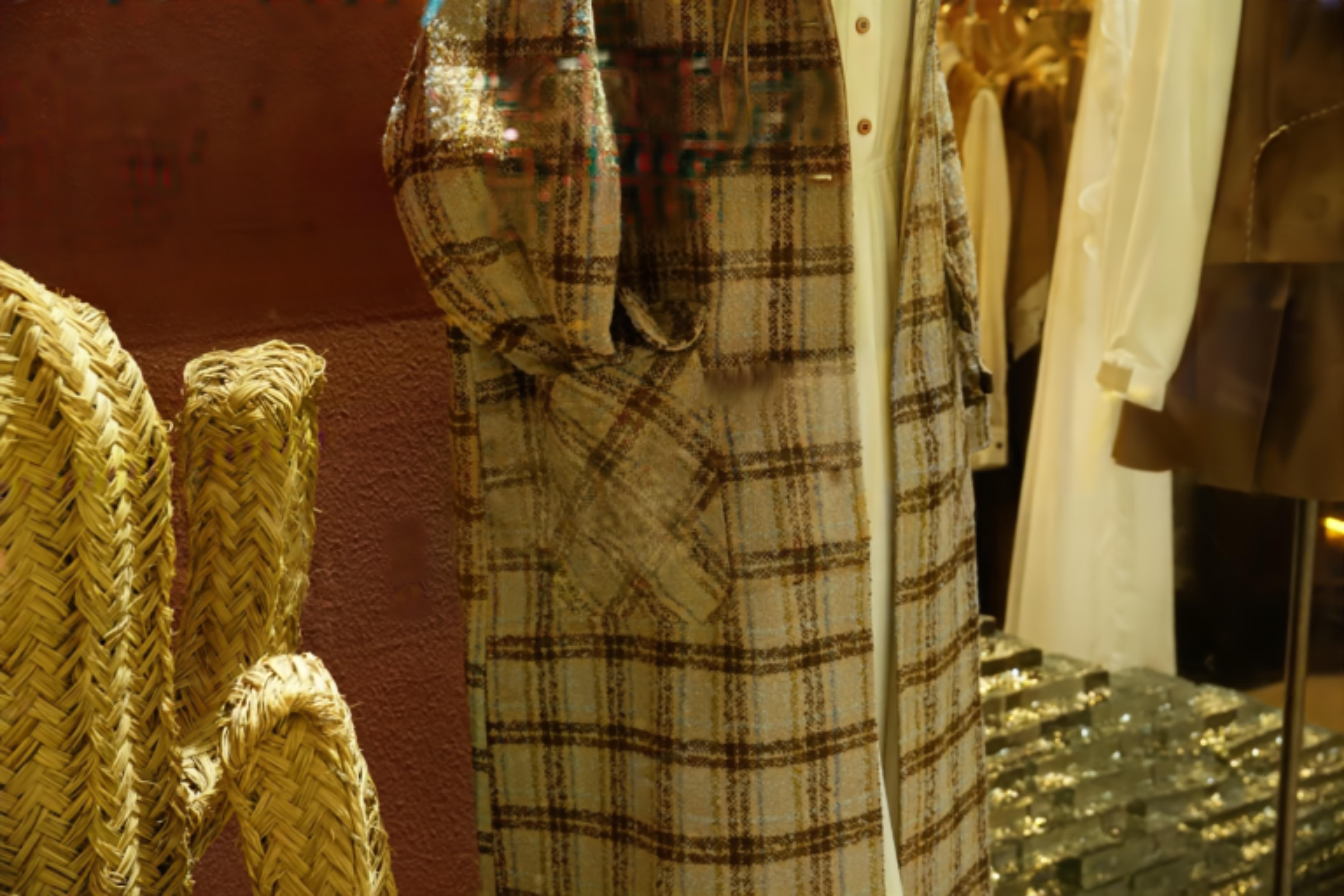}
    \includegraphics[width=0.24\textwidth]{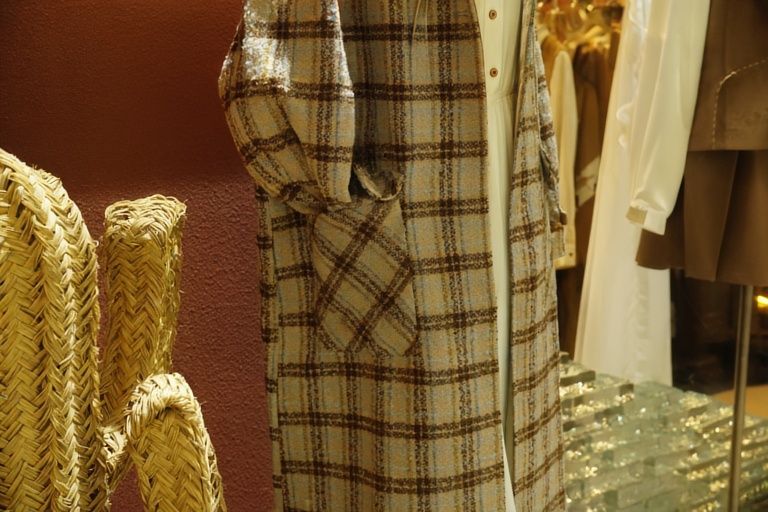}
    
    \includegraphics[width=0.24\textwidth]{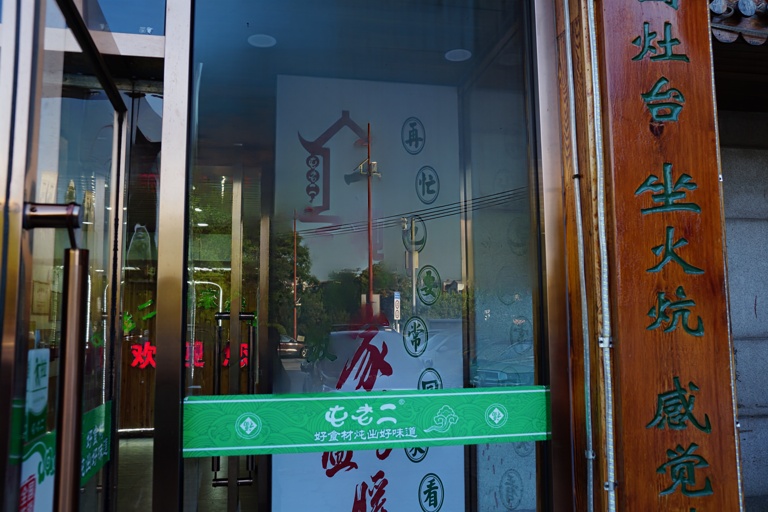}
    \includegraphics[width=0.24\textwidth]{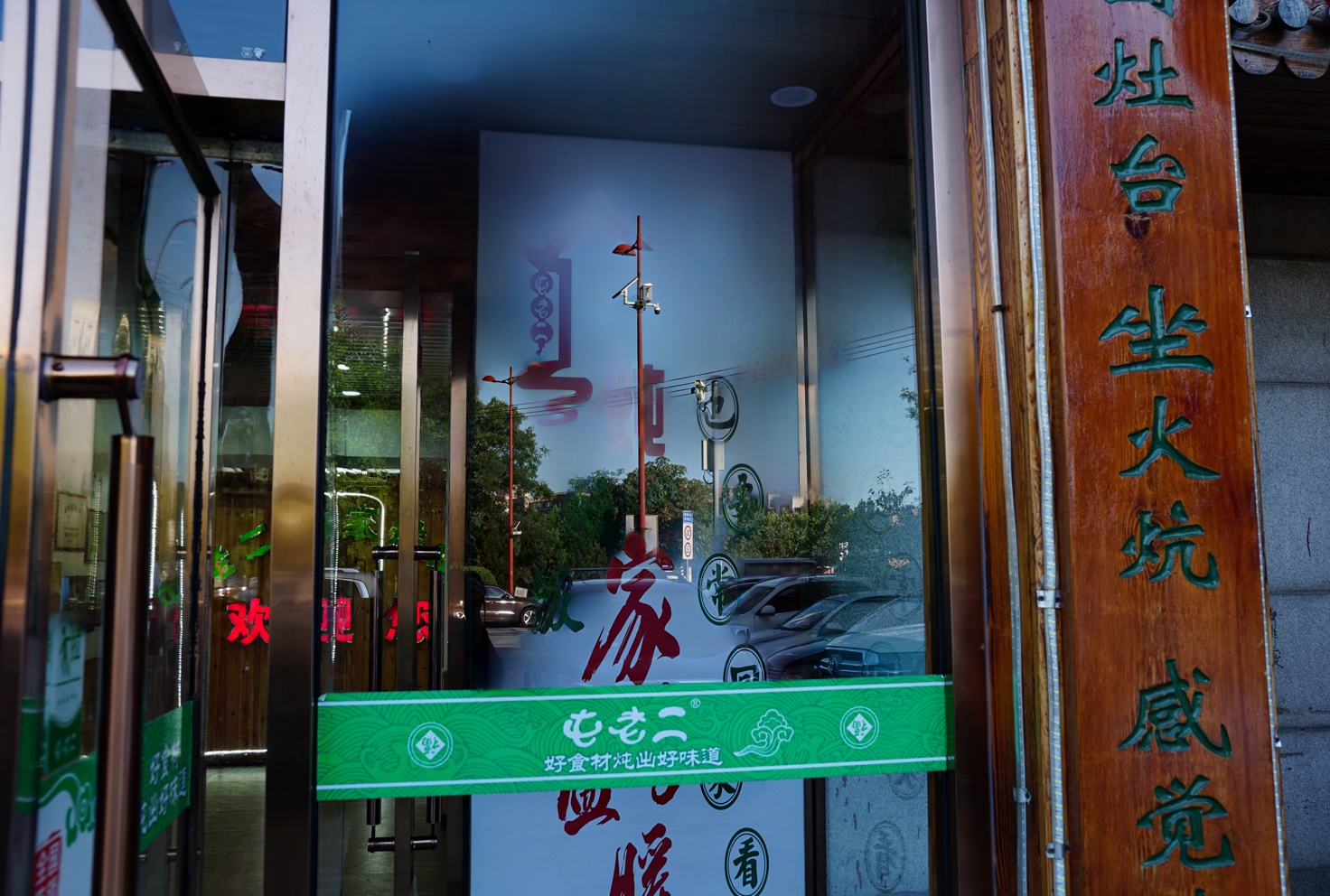}
    \includegraphics[width=0.24\textwidth]{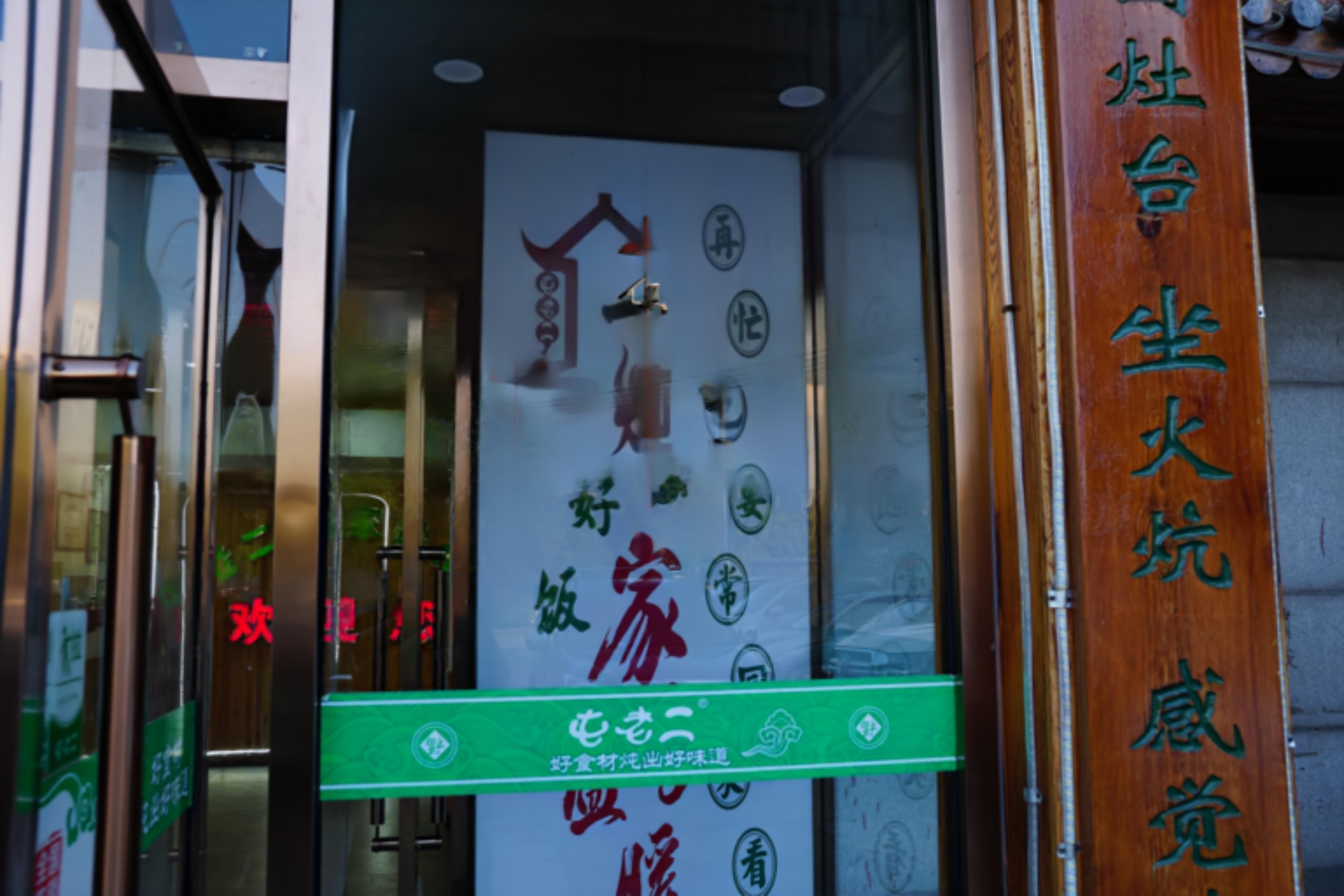}
    \includegraphics[width=0.24\textwidth]{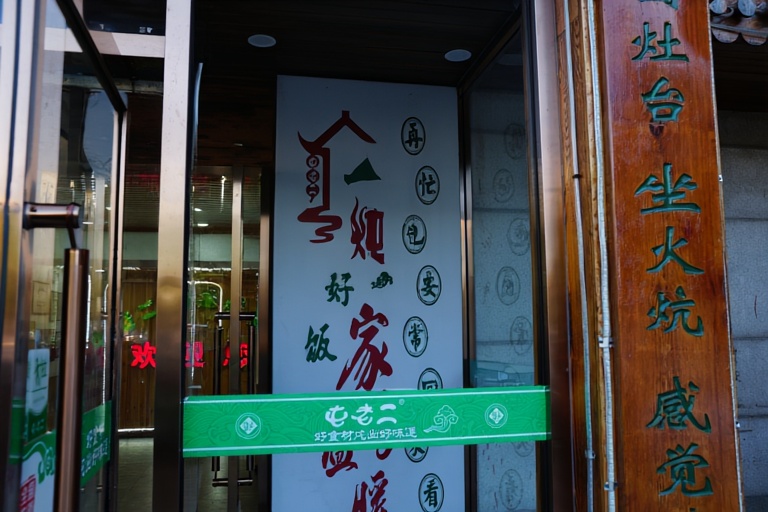}
    
   \begin{subfigure}{0.24\textwidth}
        \centering
        \subcaption{DSIT}
        \label{subfig:YTMT}
    \end{subfigure}
    \begin{subfigure}{0.24\textwidth}
        \centering
        \subcaption{RDNet}
        \label{subfig:DSRNet}
    \end{subfigure}
    \begin{subfigure}{0.24\textwidth}
        \centering
        \subcaption{DAI}
        \label{subfig:zhu}
    \end{subfigure}
    \begin{subfigure}{0.24\textwidth}
        \centering
        \subcaption{Ours}
        \label{subfig:input}
    \end{subfigure}
    \caption{Qualitative comparisons on real-world cases. Please zoom in for more details.}
    \label{fig:real}
\end{figure*}

\subsection{Qualitative Performance Evaluation}

\noindent\textit{Visual Results.} We provide extensive visual comparisons in Figure~\ref{fig:dataset} to highlight the qualitative superiority of our method. Across a variety of challenging real-world examples, prior state-of-the-art approaches exhibit common failure modes: they either leave behind noticeable residual reflection artifacts (ghosting), or they over-smooth the image, destroying fine details and textures in the underlying transmission layer. In contrast, our method is visibly more effective at removing complex reflections while preserving scene fidelity compared to prior state-of-the-art approaches. 

\noindent\textit{Generative Ability.} We compare GenSIRR with another method that utilizes a large-scale pre-trained diffusion model, DAI~\citep{DBLP:journals/corr/abs-2503-17347}, in Figure~\ref{fig:real_occu}. Under these occluded scenarios, DAI~\citep{DBLP:journals/corr/abs-2503-17347} defaults to producing a blurry, non-semantic output, effectively averaging the colors in the contaminated area. In contrast, our GenSIRR can generate reasonable texture. This confirms that our GenSIRR is indeed a generative model that outputs new textual information.
\begin{figure}[t]
    \centering 
    \includegraphics[width=\linewidth]{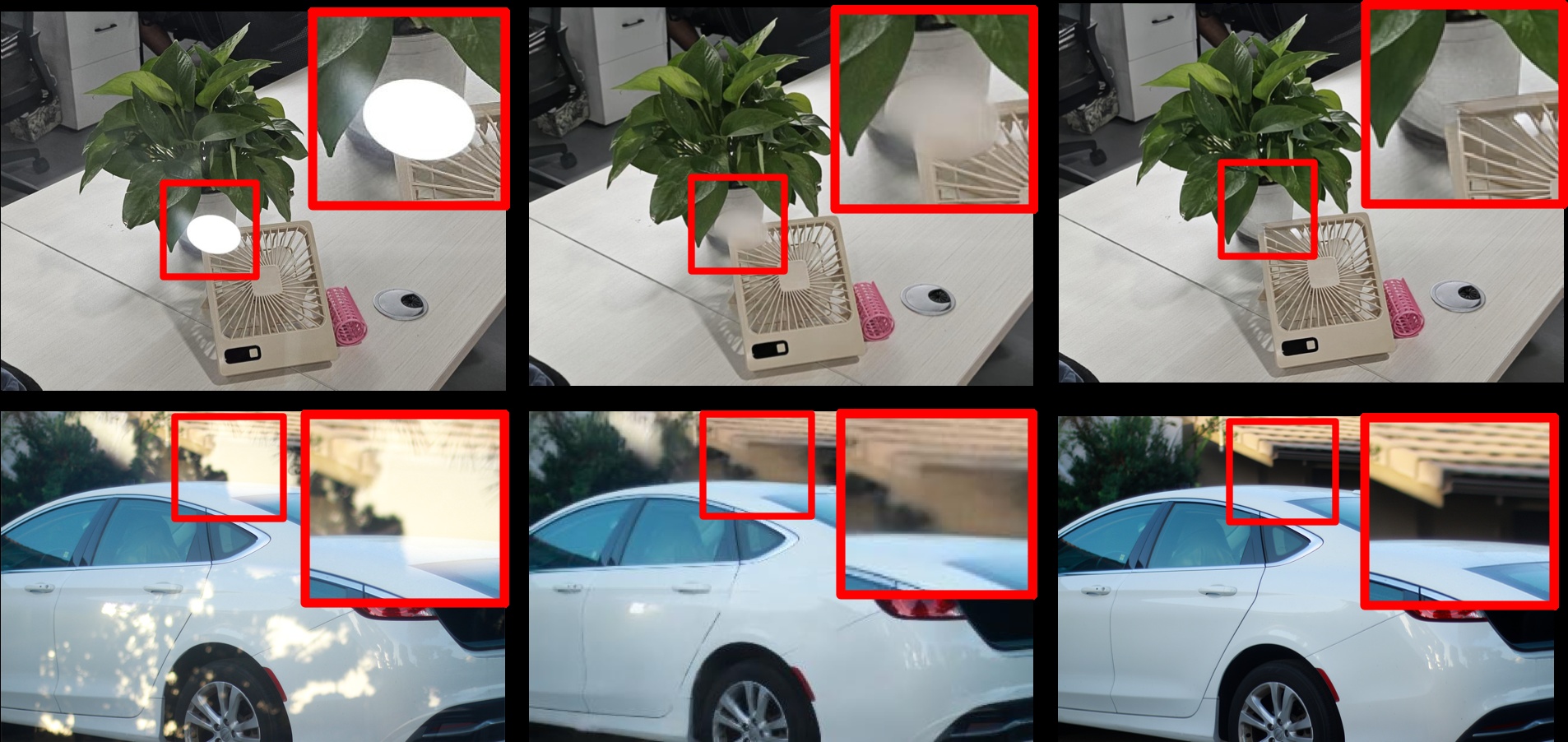}
   
   \begin{subfigure}{0.32\linewidth}
        \centering
        \subcaption{Input}
        \label{subfig:input}
    \end{subfigure}
    \begin{subfigure}{0.32\linewidth}
        \centering
        \subcaption{DAI}
        \label{subfig:DAI}
    \end{subfigure}
    \begin{subfigure}{0.32\linewidth}
        \centering
        \subcaption{Ours}
        \label{subfig:Ours}
    \end{subfigure}
    \caption{Qualitative comparisons on highly occluded cases.}
    \label{fig:real_occu}
\end{figure}

\noindent\textit{Generalization to Challenge In-the-Wild Images.} The strength of GenSIRR is most evident in its performance on real-world images, as in Figure~\ref{fig:real}. While competing methods often fail to adapt to the domain shift, producing unnatural artifacts or incomplete removal of reflections, our model demonstrates remarkable robustness. GenSIRR effectively removes complex reflections while preserving the natural appearance and details of the underlying scene, confirming its superior generalization capabilities.

\begin{table*}[t]
    \centering
    \begin{minipage}{0.33\textwidth}
        \centering
        \setlength{\tabcolsep}{4pt}
        \setlength{\tabcolsep}{2.4pt}
        \begin{tabular}{cccccc}
            \toprule
            Choice & w/o DEBS & 4 & 8 & 16 \\
            \midrule
            PSNR &  27.27 & 27.58  &27.72  &27.74 \\
            SSIM &  0.871   & 0.881   &0.879 & 0.882 \\
            \bottomrule
        \end{tabular}
        \vspace*{-0.5\baselineskip}
        \captionof{table}{The choice of $k$.}
           \label{tab:ablation_k}
    \end{minipage}
    \hfill
    \begin{minipage}{0.32\textwidth}
        \centering
        \setlength{\tabcolsep}{4pt}

        \setlength{\tabcolsep}{2.4pt}
        \begin{tabular}{cccc}
            \toprule
            Choice & w/o Training & $\mathcal{L}_{\mathrm{recon}}$ & Ours \\
            \midrule
            PSNR &  25.72   &  25.79   & 27.27 \\
            SSIM &  0.841    &  0.842    & 0.871 \\
            \bottomrule
        \end{tabular}
        \vspace*{-0.5\baselineskip}
        \captionof{table}{The choice of training VAE.}
        \label{tab:ablation_vae}
    \end{minipage}
    \hfill 
    \begin{minipage}{0.32\textwidth}
        \centering
        \setlength{\tabcolsep}{4pt}
        \setlength{\tabcolsep}{2.4pt} 
        \begin{tabular}{ccccc}
            \toprule
            Setup & Rand & RDNet  & Fix & Ours \\
            \midrule
            PSNR & N/A     &  N/A &  26.52 & 27.27 \\
            SSIM & N/A     &  N/A &  0.830   & 0.871 \\
            \bottomrule
        \end{tabular}
        \vspace*{-0.5\baselineskip}
        \captionof{table}{The choice of prompt.}
        \label{tab:ablation_prompt}
    \end{minipage}
\end{table*}

\begin{table*}[t!]
\centering
\setlength{\tabcolsep}{1pt} 
\begin{tabular}{@{}l ccc c ccc c ccc c ccc@{}}
\toprule
& \multicolumn{3}{c}{{OpenRR-val}} & \phantom{a} & \multicolumn{3}{c}{{Nature}} & \phantom{a} & \multicolumn{3}{c}{{Real20}} & \phantom{a} & \multicolumn{3}{c}{{SIR2}} \\
\cmidrule{2-4} \cmidrule{6-8} \cmidrule{10-12} \cmidrule{14-16}
{Metric} & {RDNet} & {DAI} & {Ours} && {RDNet} & {DAI} & {Ours} && {RDNet} & {DAI} & {Ours} && {RDNet} & {DAI} & {Ours} \\
\midrule
\multicolumn{16}{l}{\textit{Aggregated Success Metrics}} \\
\quad Average Success Rate & 30.4\% & 41.2\% & \textbf{96.6\%} && 63.0\% & 76.0\% & \textbf{96.0\%} && 34.0\% & 33.0\% & \textbf{91.0\%} && 16.5\% & 34.2\% & \textbf{78.5\%} \\
\quad Consensus (All Users) & 12.0\% & 20.0\% & \textbf{87.0\%} && 55.0\% & 55.0\% & \textbf{90.0\%} && 5.0\% & 10.0\% & \textbf{80.0\%} && 2.2\% & 9.9\% & \textbf{56.4\%} \\
\quad At Least One User & 52.0\% & 62.0\% & \textbf{100.0\%} && 75.0\% & 90.0\% & \textbf{100.0\%} && 60.0\% & 65.0\% & \textbf{100.0\%} && 40.1\% & 63.4\% & \textbf{93.2\%} \\
\midrule
\multicolumn{16}{l}{\textit{Per-User Success Rates}} \\
\quad User 1 & 38.0\% & 50.0\% & \textbf{97.0\%} && 55.0\% & 80.0\% & \textbf{95.0\%} && 50.0\% & 40.0\% & \textbf{85.0\%} && 20.5\% & 42.7\% & \textbf{85.5\%} \\
\quad User 2 & 25.0\% & 37.0\% & \textbf{93.0\%} && 70.0\% & 80.0\% & \textbf{95.0\%} && 35.0\% & 25.0\% & \textbf{90.0\%} && 14.5\% & 35.5\% & \textbf{77.3\%} \\
\quad User 3 & 31.0\% & 41.0\% & \textbf{95.0\%} && 65.0\% & 60.0\% & \textbf{95.0\%} && 30.0\% & 25.0\% & \textbf{95.0\%} && 6.2\% & 17.8\% & \textbf{69.6\%} \\
\quad User 4 & 32.0\% & 44.0\% & \textbf{98.0\%} && 60.0\% & 70.0\% & \textbf{95.0\%} && 25.0\% & 35.0\% & \textbf{95.0\%} && 22.9\% & 34.1\% & \textbf{80.0\%} \\
\quad User 5 & 26.0\% & 34.0\% & \textbf{100.0\%} && 65.0\% & 90.0\% & \textbf{100.0\%} && 30.0\% & 40.0\% & \textbf{90.0\%} && 18.3\% & 40.7\% & \textbf{80.4\%} \\
\midrule
\multicolumn{16}{l}{\textit{Failure Mode Analysis (Average)}} \\
\quad Incomplete Removal & 62.4\% & 54.0\% & \textbf{2.2\%} && 33.0\% & 19.0\% & \textbf{4.0\%} && 57.0\% & 57.0\% & \textbf{6.0\%} && 79.7\% & 61.6\% & \textbf{18.5\%} \\
\quad Content Deletion & 2.2\% & 1.2\% & \textbf{0.8\%} && 1.0\% & 3.0\% & \textbf{0.0\%} && 1.0\% & 5.0\% & \textbf{0.0\%} && 1.4\% & \textbf{1.3}\% & 2.2\% \\
\quad Artifact Generation & 5.0\% & 3.6\% & \textbf{0.4\%} && 3.0\% & 2.0\% & \textbf{0.0\%} && 8.0\% & 5.0\% & \textbf{3.0\%} && 2.4\% & 2.9\% & \textbf{0.7\%} \\
\bottomrule
\end{tabular}
\caption{Comprehensive human evaluation results across four benchmark datasets: \textit{OpenRR-val}, \textit{Nature}, \textit{Real20}, and \textit{SIR2}. We report aggregate success metrics, individual evaluator scores, and a breakdown of failure modes. ``Consensus'' indicates the percentage of samples where \textit{all} evaluators agreed on success. Best results are in \textbf{bold}.}
\label{tab:user_study_all_datasets}
\end{table*}

\subsection{Human Evaluation}
To complement the quantitative metrics, we conducted a rigorous human evaluation across four benchmark datasets: OpenRR-val, Nature, Real20, and SIR2. Unlike subjective preference studies, this evaluation is objective: 5 evaluators were asked to assess whether the reflection was successfully removed by comparing the output against the ground truth. The outputs were classified into four categories:
\begin{itemize}
    \item \textbf{Success.} Reflection is cleanly removed, preserving scene fidelity without artifacts.
    \item \textbf{Failure (Incomplete).}Obvious  reflections remain.
    \item \textbf{Failure (Content Deletion).}  Important background details are removed or distorted.
    \item \textbf{Failure (Artifacts).} The model introduces unrealistic patterns or geometric distortions.
\end{itemize}

The results, detailed in Table~\ref{tab:user_study_all_datasets}, demonstrate the robustness of our method. On the OpenRR-val dataset, our method achieves a dominant average success rate of 96.6\%, significantly outperforming DAI (41.2\%) and RDNet (30.4\%). Notably, we achieved a consensus success rate (where all 5 evaluators agreed) of 87.0\%, and in 100\% of samples, at least one evaluator marked the result as successful. This trend holds across all datasets, with ours consistently achieving success rates above 78\% even on the challenging SIR2 dataset, while competitors struggle with incomplete removal rates as high as 79\%. We attribute the lower success rates of competing methods to their reliance on pixel-wise reconstruction objectives (e.g., MSE/L1), even in the case of one-step diffusion fine-tuning. These objectives often encourage models to merely suppress or dim reflections to minimize statistical error rather than structurally separating the layers. This results in visible residual that evaluators strictly penalize.

\subsection{Ablation Analysis}
To validate the component's effectiveness in our GenSIRR framework, we conduct ablations on the Real20 dataset. The quantitative results are in Tables~\ref{tab:ablation_k}, ~\ref{tab:ablation_vae},and \ref{tab:ablation_prompt}. 

\noindent\textit{On the Rationale of the re-VAE.}
We hypothesize that a structured latent space is helpful for SIRR. To verify this, we train a variant of our model where the reflection-equivariant VAE is replaced with a standard VAE trained only with a reconstruction loss ($\mathcal{L}_{\text{recon}}$), removing the linearity constraint ($\mathcal{L}_{\text{equiv}}$). As shown in Table~\ref{tab:ablation_vae}, both changes lead to a significant drop in performance. This experiment confirms that the reflection-equivalence enables a precise and complete removal of the reflection.

\noindent\textit{On the Efficacy of LTE.}
We argue that learnable prompts provide a more precise and effective form of task guidance than fixed text. To ablate this component, we replace our learnable prompts with a fixed text prompt, "Please remove the reflection in this image." The results in Table~\ref{tab:ablation_prompt} show a marked degradation in performance. In some cases, it fails to remove complex reflection patterns. We also tried to randomly initialize the prompt or use RDNet~\citep{cvpr/zhao2025reversible}'s prompt generator to generate a prompt. However, both choices failed to generate meaningful content, and the loss doesn't converge. A possible reason may involve the vast nature of the prompt embedding space. A randomly initialized prompt lacks any semantic anchor, forcing the model to search an enormous space for a meaningful task vector from scratch. Without an initial direction, the optimization process fails to find a useful signal. Similarly, while the RDNet prompt generator is effective within its own architecture, its output is not semantically aligned with the pre-trained knowledge of models like FLUX.1, and thus also fails to provide a valid starting point.

\noindent\textit{On the Choice of $k$ in DEBS.} We investigate the impact of the number of parallel samples, $k$, in our DEBS strategy. This governs the breadth of exploration. We conducted an ablation study on the Real20 dataset by varying $k \in \{4, 8, 16\}$, with the results in Table~\ref{tab:ablation_k}. The performance, measured in both PSNR and SSIM, steadily improves once DEBS is enabled and increases as $k$ increases from 4 to 16. This demonstrates the effectiveness of DEBS. By exploring multiple initial trajectories, DEBS can more reliably identify a path that successfully removes reflections compared to a single, deterministic run (w/o DEBS). However, we found that the performance gains saturate when increasing the sample count further to $k=8$. This indicates that with $8$ candidates, our depth-guided scoring is already highly likely to select a high-quality trajectory, and additional sampling yields diminishing returns. Table~\ref{tab:debs_complexity} presents the inference time and memory usage for varying candidate counts ($k \in \{4, 8, 16\}$) on $256\times256$ inputs. Even with $k=16$, the increase in total inference time is marginal compared to the significant gain in stability and success rate.

\begin{table}[t!]
\centering
\setlength{\tabcolsep}{6pt}
\begin{tabular}{lccc}
\toprule
{Configuration} & {Time (ms)}  & {Relative Cost} \\
\midrule
Baseline ($k=1$) & 2061.9 & 1.00$\times$ \\
DEBS ($k=4$)  & 2580.0 & 1.25$\times$ \\
DEBS ($k=8$)  & 3005.8 & 1.46$\times$ \\
DEBS ($k=16$) & 3952.4 & 1.92$\times$ \\
\bottomrule
\end{tabular}
\caption{Computational cost of DEBS with varying candidate counts ($k$) compared to the baseline. We report the total inference time (ms) for generating a $256\times256$ image. The overhead is low because the multi-path exploration is restricted to the 1st timestep.}
\label{tab:debs_complexity}
\end{table}

\section{Limitations and Future Work}

The primary limitation of GenSIRR lies in its inference latency. Built upon an iterative diffusion process with a large transformer backbone, our framework requires multiple reverse sampling steps (28 steps in our work), resulting in an inference time of approximately 2 seconds per image on 256 $\times$ 256 images. This contrasts with prior non-generative methods, such as RDNet, and one-step fine-tuning of diffusive pretraining, like DAI, which operate in under 100ms via a single forward pass. However, as shown in Table~\ref{tab:speed_benchmark}, this computational cost yields a decisive advantage in reliability: while faster methods fail to remove reflections in the majority of cases (success rates of 30.4\% and 41.2\%), our method achieves a \textit{90.5\%} success rate. We argue that for photography and restoration tasks, a few seconds of processing time is an acceptable trade-off for a result that is perceptually successful and free of artifacts.

Nevertheless, mitigating this latency remains a priority. Potential strategies include model distillation to transfer knowledge to a lightweight student network or the adoption of few-step sampling techniques to reduce iteration counts without compromising quality. Investigating these acceleration strategies will be a key step toward broadening the applicability of our high-fidelity reflection removal.

\begin{table}[t!]
\centering
\setlength{\tabcolsep}{3pt}
\begin{tabular}{lccc}
\toprule
{Method} & {Time (ms)} & {S. Rate (\%)} & {Avg. PSNR (dB)} \\
\midrule
RDNet & \textbf{90.1} & 36.0 & 26.63 \\
DAI & 97.2 & 46.1 & 27.35 \\
{Ours} & 2580.0 & \textbf{90.5} & \textbf{28.03} \\
\bottomrule
\end{tabular}
\caption{Comparison of inference speed and performance. We report inference time (ms) on a single NVIDIA A800 GPU ($256\times256$), alongside the user study success rate (S. Rate) and Average PSNR. While our method is slower, the dramatic increase in success rate justifies the computation. Best results are in \textbf{bold}.}
\label{tab:speed_benchmark}
\end{table}

\section{Conclusion}
In this work, we introduced a novel pipeline for  SIRR. We first identified a critical yet overlooked issue, \textit{i.e.}, pre-trained generative models fail to comprehend the linear superposition nature of reflection-corrupted images within their latent space. To overcome this issue, our paradigm transforms a general-purpose generative model into a precise SIRR tool via integrating a \emph{reflection-equivariant VAE} for structuring the latent space to align with the physics of reflection formation, a \emph{learnable task-specific text embedding} that provides direct, optimized task guidance, and a \emph{depth-guided early-branching sampling strategy} that mitigates generative stochasticity by selecting the most physically plausible result. Extensive experimental results have been conducted to verify the effectiveness of our method by advanced performance on public benchmarks, and reveal its exceptional robustness on challenging real-world images. 
\section*{Acknowledgment}
This work was supported by the National Natural Science Foundation of China under Grant
No. 62372251. The computational resources of this work is partially supported by TPU Research Cloud
(TRC). 

{
    \small
    \bibliographystyle{ieeenat_fullname}
    \bibliography{main}
}

\end{document}